%% file: main.tex
\documentclass[11pt, a4paper]{lumia}

\usepackage[sort&compress]{natbib}
\input{notations}

\usepackage{graphicx}           
\usepackage{tikz}               
\usepackage[edges]{forest}      

\usepackage{url}                
\usepackage{xurl}               

\usepackage{array}              
\usepackage{longtable}          
\usepackage{multirow}           
\usepackage{makecell}           
\usepackage{ragged2e}           

\usepackage{mathtools}          
\usepackage{nicefrac}           

\usepackage{algorithm}          
\usepackage{algorithmicx}       
\usepackage{algpseudocode}      
\usepackage{listings}           

\usepackage{subcaption}         
\usepackage{wrapfig}            
\usepackage[export]{adjustbox}  

\usepackage{changes}            
\usepackage{xspace}             
\usepackage[normalem]{ulem}     
\usepackage{CJKutf8}            

\usepackage[tikz]{bclogo}       
\usepackage[framemethod=tikz]{mdframed} 

\usepackage{lipsum}             
\usepackage{tocloft}            
\usepackage{afterpage}          
\usepackage{bbding}             
\usepackage{epigraph}           
\usepackage{minitoc}            
\usepackage{multicol}           
\usepackage{textgreek}          

\newcolumntype{P}[1]{>{\RaggedRight\arraybackslash}p{#1}}

\definecolor{uclablue}{RGB}{39, 116, 174}
\definecolor{bigaired}{RGB}{156, 0, 0}
\definecolor{myblue}{HTML}{598BE7}
\definecolor{mildblue}{RGB}{31,119,180}
\definecolor{sectionblue}{RGB}{70, 130, 180}
\definecolor{methodblue}{RGB}{0, 150, 136}
\definecolor{bgblue}{RGB}{245,243,253}
\definecolor{ttblue}{RGB}{91,194,224}
\definecolor{mygreen}{rgb}{0.64, 0.56, 0.88}
\definecolor{myyellow}{rgb}{0.68, 0.6, 0.1}
\definecolor{fancygreen}{rgb}{0.33, 0.68, 0.20}
\definecolor{salmon}{rgb}{0.94, 0.52, 0.49}
\definecolor{tablegreen}{rgb}{0.82, 0.94, 0.75}
\definecolor{tableblue}{rgb}{0.81, 0.90, 0.94}
\definecolor{tablered}{rgb}{0.97, 0.85, 0.85}
\definecolor{tableorange}{rgb}{0.96, 0.85, 0.81}
\definecolor{myorange}{rgb}{1.0, 0.49, 0.0}
\definecolor{tlgreen}{rgb}{0.33, 0.68, 0.20}
\definecolor{darkgreen}{RGB}{0,100,0}
\definecolor{darkred}{RGB}{200, 0, 0}
\definecolor{customyellow}{HTML}{FFFACD}
\definecolor{refinegreen}{RGB}{0, 128, 75}
\definecolor{scoregreen}{RGB}{34, 139, 34}
\definecolor{hidden-blue}{RGB}{194,232,247}
\definecolor{hidden-black}{RGB}{20,68,106}
\definecolor{yes}{HTML}{C6EFCE}
\definecolor{no}{HTML}{FFC7CE}
\definecolor{partial}{HTML}{FFEB9C}
\definecolor{external}{HTML}{D9E1F2}
\definecolor{hdr}{HTML}{F2F2F2}
\definecolor{GRPOrow}{gray}{0.96}
\definecolor{FlowRLrow}{RGB}{225,236,255}
\definecolor{FlowBlue}{RGB}{80,120,210}
\definecolor{GRPOGray}{gray}{0.35}

\setlist[itemize]{leftmargin=20pt, noitemsep, topsep=0pt}

\NewDocumentCommand{\kaiyan}{mO{}}{\textcolor{purple}{\textsuperscript{\textit{kaiyan}}\textsf{\textbf{\small[#1]}}}}
\NewDocumentCommand{\yuxin}{mO{}}{\textcolor{cyan}{\textsuperscript{\textit{yuxin}}\textsf{\textbf{\small[#1]}}}}
\NewDocumentCommand{\bx}{mO{}}{\textcolor{green}{\textsuperscript{\textit{bx}}\textsf{\textbf{\small[#1]}}}}
\NewDocumentCommand{\at}{mO{}}{\textcolor{red}{\textsuperscript{\textit{AT}}\textsf{\textbf{\small[#1]}}}}
\NewDocumentCommand{\re}{mO{}}{\textcolor{blue}{\textsuperscript{\textit{RE}}\textsf{\textbf{\small[#1]}}}}
\NewDocumentCommand{\ybsun}{mO{}}{\textcolor{magenta}{\textsuperscript{\textit{youbang}}\textsf{\textbf{\small[#1]}}}}
\NewDocumentCommand{\runze}{mO{}}{\textcolor{orange}{\textsuperscript{\textit{runze}}\textsf{\textbf{\small[#1]}}}}
\NewDocumentCommand{\add}{mO{}}{\textcolor{darkgreen}{\textsuperscript{\textit{Maybe Consider Discuss}}\textsf{\textbf{[#1]}}}}

\newcommand{\cmark}{\textcolor{darkgreen}{\boldmath$\checkmark$}}
\newcommand{\xmark}{\textcolor{darkred}{\boldmath$\times$}}

\newenvironment{itemize*}%
 {\leftmargini=10pt\begin{itemize}%
  \setlength{\itemsep}{0pt}%
  \setlength{\parskip}{0pt}%
  }%
 {\end{itemize}}

\newenvironment{enumerate*}%
 {\begin{enumerate}%
  \setlength{\itemsep}{0pt}%
  \setlength{\parskip}{0pt}}%
 {\end{enumerate}}

\newcommand{\cellstatus}[1]{%
  \begingroup
  \StrTrim{#1}[\statusval]%
  \IfStrEq{\statusval}{Yes}{\cellcolor{yes}\cmark}{}%
  \IfStrEq{\statusval}{No}{\cellcolor{no}\xmark}{}%
  \IfBeginWith{\statusval}{Yes (}{\cellcolor{yes}\cmark~\textit{\statusval\unskip}}{}%
  \IfStrEq{\statusval}{Partial}{\cellcolor{partial}\textbf{Partial}}{}%
  \IfStrEq{\statusval}{External}{\cellcolor{external}\textbf{External}}{}%
  \endgroup
}

\newtcolorbox{myboxi}[1][]{
  breakable,
  title=#1,
  colback=red!5,
  colbacktitle=red!5,
  coltitle=black,
  fonttitle=\bfseries,
  bottomrule=0pt,
  toprule=0pt,
  leftrule=2pt,
  rightrule=2pt,
  titlerule=0pt,
  arc=0pt,
  outer arc=0pt,
  colframe=red,
}

\newtcolorbox{myboxnote}[1][]{
  breakable,
  title=#1,
  colback=orange!0,
  colbacktitle=orange!0,
  coltitle=black,
  fonttitle=\bfseries,
  bottomrule=0pt,
  toprule=0pt,
  leftrule=2pt,
  rightrule=2pt,
  titlerule=0pt,
  arc=0pt,
  outer arc=0pt,
  colframe=orange,
}

\newtcolorbox{myboxii}[1][]{
  breakable,
  freelance,
  title=#1,
  colback=white,
  colbacktitle=white,
  coltitle=black,
  fonttitle=\bfseries,
  bottomrule=0pt,
  boxrule=0pt,
  colframe=white,
  overlay unbroken and first={
  \draw[red!75!black,line width=3pt]
    ([xshift=5pt]frame.north west) -- 
    (frame.north west) -- 
    (frame.south west);
  \draw[red!75!black,line width=3pt]
    ([xshift=-5pt]frame.north east) -- 
    (frame.north east) -- 
    (frame.south east);
  },
  overlay unbroken app={
  \draw[red!75!black,line width=3pt,line cap=rect]
    (frame.south west) -- 
    ([xshift=5pt]frame.south west);
  \draw[red!75!black,line width=3pt,line cap=rect]
    (frame.south east) -- 
    ([xshift=-5pt]frame.south east);
  },
  overlay middle and last={
  \draw[red!75!black,line width=3pt]
    (frame.north west) -- 
    (frame.south west);
  \draw[red!75!black,line width=3pt]
    (frame.north east) -- 
    (frame.south east);
  },
  overlay last app={
  \draw[red!75!black,line width=3pt,line cap=rect]
    (frame.south west) --
    ([xshift=5pt]frame.south west);
  \draw[red!75!black,line width=3pt,line cap=rect]
    (frame.south east) --
    ([xshift=-5pt]frame.south east);
  },
}

\tcbset{
  takeawaysbox/.style={
    title=Takeaways,
    colback=lightblue!80,
    colframe=black,
    fonttitle=\bfseries\small,
    coltitle=white,
    colbacktitle=black,
    enhanced,
    attach boxed title to top left={xshift=2.5mm,yshift=-2.5mm},
    boxed title style={rounded corners, size=small, colframe=black, colback=black},
    width=\linewidth,
    arc=3.5mm
  }
}

\mdfdefinestyle{mystyle}{%
  rightline=true,
  innerleftmargin=10,
  innerrightmargin=10,
  outerlinewidth=3pt,
  topline=false,
  rightline=true,
  bottomline=false,
  skipabove=\topsep,
  skipbelow=\topsep
}

\tikzset{%
    every node/.style={font=\tiny},
    parent/.style =          {align=center,text width=2cm,rounded corners=3pt, line width=0.3mm, fill=gray!10,draw=gray!80},
    child/.style =           {align=center,text width=2.0cm,rounded corners=3pt, fill=blue!10,draw=blue!80,line width=0.3mm},
    grandchild/.style =      {align=center,text width=2cm,rounded corners=3pt},
    greatgrandchild/.style = {align=center,text width=1.5cm,rounded corners=3pt},
    greatgrandchild2/.style = {align=center,text width=1.5cm,rounded corners=3pt},    
    referenceblock/.style =  {align=center,text width=1.5cm,rounded corners=2pt},
    pretrain/.style =           {align=center,text width=2.0cm,rounded corners=3pt, fill=blue!10,draw=blue!80,line width=0.3mm},   
    pretrain_work/.style =           {align=center, text width=8.5cm,rounded corners=3pt, fill=blue!10,draw=blue!0,line width=0.3mm},  
    template/.style =           {align=center,text width=2.0cm,rounded corners=3pt, fill=red!10,draw=red!80,line width=0.3mm},   
    template_work/.style =           {align=center,text width=8.5cm,rounded corners=3pt, fill=red!10,draw=red!0,line width=0.3mm},    
    answer/.style =           {align=center,text width=2.0cm,rounded corners=3pt, fill= cyan!10,draw= cyan!80,line width=0.3mm},   
    answer_work/.style =           {align=center,text width=8.5cm,rounded corners=3pt, fill= cyan!10,draw= cyan!0,line width=0.3mm},      
    multiple/.style =           {align=center,text width=2.0cm,rounded corners=3pt, fill= orange!10,draw= orange!80,line width=0.3mm},   
    multiple_work/.style =           {align=center,text width=8.5cm,rounded corners=3pt, fill= orange!10,draw= orange!0,line width=0.3mm},        
    tuning/.style =           {align=center,text width=2.0cm,rounded corners=3pt, fill= magenta!10,draw= magenta!80,line width=0.3mm},   
    tuning_work/.style =           {align=center,text width=8.5cm,rounded corners=3pt, fill= magenta!10,draw= magenta!0,line width=0.3mm},          
}

\newcommand{\lstbg}[3][0pt]{{\fboxsep#1\colorbox{#2}{\strut #3}}}

\lstdefinelanguage{diff}{
  basicstyle=\ttfamily\small,
  morecomment=[f][\lstbg{red!20}]-,
  morecomment=[f][\lstbg{green!20}]+,
}

\lstdefinelanguage{diffpython}{
  language=diff,
  morekeywords={def, if, else, for, while, return, import, from, as, class, with, try, except, finally, raise, lambda, and, or, not, in, is, None, True, False},
  morecomment=[l]{\#},
  morestring=[b]",
  morestring=[b]',
}

\usepackage{xspace}

\usepackage{xcolor}

\definecolor{ForestGreen}{RGB}{34,139,34}
\definecolor{myyellow}{RGB}{181, 181, 27}

\usepackage[table,xcdraw,usenames,dvipsnames]{xcolor}

\usepackage{amsmath}
\usepackage{amssymb}
\usepackage{mathtools}
\usepackage{amsthm}
\usepackage{fontawesome5} 

\usepackage[capitalize,noabbrev]{cleveref}

\usepackage{multirow}
\usepackage{makecell}
\usepackage{enumerate}
\usepackage{enumitem}
\usepackage{pifont}

\definecolor{mygrey}{gray}{0.4}

\usepackage[ruled,algo2e,vlined]{algorithm2e}
\SetKwInOut{Input}{Input}\SetKwInOut{Output}{Output}
\SetKwComment{Comment}{$\triangleright$\ }{}

\definecolor{darkgreen}{RGB}{30, 130, 30}
\definecolor{cream}{RGB}{253, 250, 242}
\renewcommand{\cmark}{\textcolor{darkgreen}{\ding{51}}} 
\renewcommand{\xmark}{\textcolor{red}{\ding{55}}}       

\usepackage{listings}

\usepackage{ulem}
\usepackage{pifont}
\def\method{{\fontfamily{lmtt}\selectfont Struct-Searcher}}

\setheadertext{LUMIA Lab}

\correspondingemail{\emailicon\ fzhang25@cse.cuhk.edu.hk \quad $*$ Equal Contribution \quad $^\ddagger$ Corresponding Author}

\title{Struct-Searcher: Agentic Structural Thinking Advances Multimodal Deep Information Seeking}
\setheadertitle{Struct-Searcher: Agentic Structural Thinking Advances Multimodal Deep Information Seeking}

\author{%
  Fan Zhang$^{1,*}$, Vireo Zhang$^{*}$, Shengju Qian$^{2\dagger}$, Haoxuan Li$^{3}$, Zheng Lian$^{4}$,\linebreak Hao Wu$^{5}$, Yuan Gao$^{5}$, Xinyu Geng$^{6}$, Xin Wang$^{2}$, Pheng-Ann Heng$^{1}$\\
  $^1$CUHK
  $^2$LIGHTSPEED
  $^3$PKU
  $^4$Tongji University
  $^5$THU
  $^6$HKUST
  \\
  
}

\begin{document}

\begin{abstract}

Deep research agents have attracted increasing attention for their ability to collect large-scale online information to acquire target knowledge, with recent efforts shifting from purely text-based information seeking to multimodal settings. However, existing agentic workflows are largely aligned with evidence accumulation models, which linearly aggregate evidence and lack principled mechanisms for handling contradictory information across heterogeneous modalities. Towards this end, we propose \method{}, a structural agentic workflow grounded in belief revision theory that explicitly maintains an evolving multimodal structural graph throughout the reasoning process, enabling effective conflict-aware multimodal deep information seeking. Extensive experiments across multiple benchmark datasets and backbone models demonstrate that \method{} is: (1) \textbf{plug-and-play} and \textbf{model-agnostic}, yielding an average relative accuracy improvement of 17.2\% on BrowseComp-VL across five different backbones. (2) \textbf{top-performing}, consistently outperforming state-of-the-art vision–language models (VLMs) and deep research agents, with relative accuracy improvements of 3.7\% on MM-BrowseComp, 1.5\% on HLE-VL, and 0.7\% on BrowseComp-VL over the second-best competing approach.
\end{abstract}

\maketitle

\begin{center}
  \resizebox{0.95\linewidth}{!}{
    \includegraphics{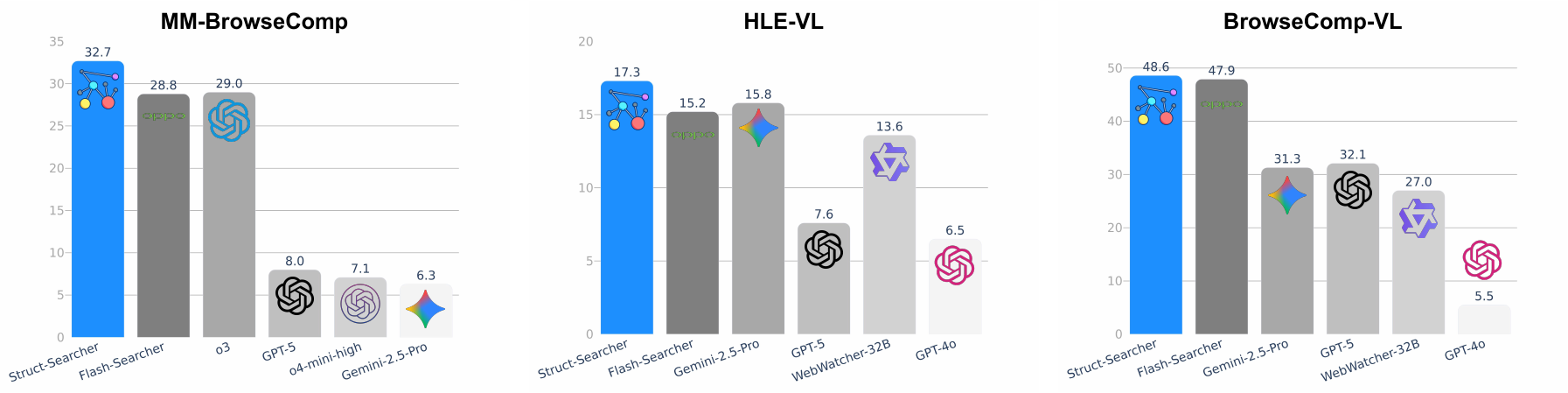}
    }
    \captionof{figure}{Pass@1 (Overall Accuracy) performance comparison ($\%$) between our \method{} and other leading industry providers across three widely used benchmarks. Both \method{} and Flash-Searcher are built on the GPT-5 backbone.
    \label{fig:intro}
	}
\end{center}


\section{Introduction}

Language agents and agentic systems, empowered by increasingly capable foundation models~\citep{kimiteam2025kimik2openagentic,meituanlongcatteam2025longcatflashtechnicalreport} and sophisticated scaffolding frameworks~\citep{openhands}, have progressed at a remarkable pace. These advances have led to substantial gains across challenging domains, including software engineering~\citep{wei2025swerladvancingllmreasoning,golubev2025traininglongcontextmultiturnsoftware} and scientific discovery~\citep{bai2025interns1scientificmultimodalfoundation,mitchener2025kosmosaiscientistautonomous}. Among the emerging directions, \emph{deep research} has attracted particular attention~\citep{zhang2025deepresearchsurveyautonomous,shi2025deepresearchsystematicsurvey}, wherein agents are required to aggregate, synthesize, and reason over large-scale online information sources with both broad coverage and considerable depth to obtain target knowledge. 
This line of work has been actively advanced by both academic efforts~\citep{li2025search,zhang2025agentorchestra} and industrial systems~\citep{perplexityIntroducingPerplexity,deepresearch,moonshotaiKimiResearcherEndtoEnd,grokGrok}, collectively driving rapid progress in this area.

Looking ahead, deep research is evolving from predominantly text-centric paradigms~\citep{song2025r1searcherincentivizingsearchcapability} toward comprehensive multimodal support~\citep{yang2025multimodaldeepresearchergeneratingtextchart}. This evolution is driven by the intrinsic complexity of real-world environments: information on the Internet is fundamentally heterogeneous and multimodal, rendering native support for \textit{images}, \textit{audio}, \textit{video}, and other modalities essential rather than optional~\citep{xu2025comprehensivesurveydeepresearch}. This claim is increasingly supported by converging evidence from both \textit{benchmarks} and \textit{frameworks}. On the benchmarking side, evaluations for deep research have progressed from early artifacts that sparsely incorporated multimodal elements~\citep{mialon2023gaia,phan2025humanity}) to more recent benchmarks that explicitly emphasize multimodal inputs and information flow~\citep{li2025mm,jiang2024mmsearchbenchmarkingpotentiallarge}). In parallel, agent frameworks are moving beyond early unimodal designs~\citep{jin2025searchr1trainingllmsreason,wu2025webdancer} and beginning to systematically embrace mixed-modality agentic pipelines.

However, the shift toward multimodal deep research is a double-edged sword. While multimodality substantially increases representational expressiveness, it also introduces severe structural challenges, including semantic ambiguity~\citep{chen2025deepresearchbringsdeeper}, temporal inconsistency across modalities~\citep{martins2025visiongeotemporaldeepresearch}, and pronounced information redundancy that complicates the rapid extraction of task-relevant evidence~\citep{tao2025mmsearchplusbenchmarkingprovenanceawaresearch}. Most critically, multimodal settings give rise to \emph{cross-modal conflict}, where evidence from different modalities contradicts rather than corroborates one another~\citep{yan2025multimodalinconsistency}. For example, a textual report may assert a product’s durability, whereas a retrieved review video exposes clear physical defects. Such discrepancies create a persistent ``truth gap,'' rendering reliability verification and logical consistency in deep research considerably more challenging than in unimodal scenarios.

One might assume that current systems could readily adapt to these demands by equipping multimodal tools or leveraging powerful multimodal reasoning models~\citep{li2025perceptionreasonthinkplan,oaio3-o4-mini} to extend their perceptual capabilities. However, we argue that such superficial integration fails to address the core epistemic challenge. Recall that the dominant paradigm in current deep research systems typically employs a hierarchical topology, where a coordinator dispatches tasks to sequential sub-agents (\textit{e.g.}, browser agent, coding agent) to gather information. Theoretically, this workflow aligns with \emph{Evidence Accumulation Models} (EAM)~\citep{evans2020evidence-eam}, where the system linearly aggregates evidence until a confidence threshold is met. While effective for consistent data streams, EAMs suffer from \emph{epistemic rigidity}: they treat evidence gathering as a monotonic process, lacking the cognitive mechanism to retract or revise prior beliefs when later encountering high-fidelity, contradictory evidence from diverse modalities. This structural limitation necessitates a fundamental paradigm shift from simple accumulation to dynamic \emph{belief revision}.

In this paper, we aim to reframe multimodal deep research systems from an \emph{evidence accumulation} paradigm to a principled \emph{belief revision} paradigm. 
Drawing inspiration from the Alchourrón--Gärdenfors--Makinson (AGM) theory~\citep{alchourron1985logic}, we introduce \method{}, a belief-driven agentic deep research framework that navigates large-scale multimodal information spaces through explicit belief construction and revision. 
Concretely, given a multimodal query, \method{} initializes a \emph{multimodal structural graph} (MSG) to represent its belief state, in which query, goal, hypothesis, and evidence nodes are explicitly encoded with their support and conflict relations. As the agent iteratively retrieves external multimodal information, it performs belief expansion, contraction, and revision over the MSG. The final answer is synthesized from a maximal conflict-free subgraph, ensuring global logical coherence. By making belief structures explicit, \method{} enables more transparent search behavior and exhibits inherent robustness to cross-modal conflicts.

Our main contributions can be summarized as follows:
\begin{itemize}[leftmargin=*]
\item[\ding{182}] \textbf{Paradigm Shift.} We reinterpret multimodal deep research systems through the lens of belief revision theory, identifying fundamental limitations of existing evidence-accumulation-based systems under cross-modal conflicts, and argue for a principled transition from linear evidence accumulation to a dynamic belief-driven paradigm.
\item[\ding{183}] \textbf{Framework Proposal.} We propose \method{}, a belief-driven multimodal deep research framework, which maintains an MSG and iteratively performs belief revision through structural thinking, synthesizing answers from maximal conflict-free subgraphs to efficiently navigate large-scale heterogeneous information spaces.
\item[\ding{184}] \textbf{Empirical Evaluation.} As previewed in Figure \ref{fig:intro}, we validate the effectiveness of \method{} through extensive experiments on three representative multimodal deep information seeking benchmarks, demonstrating its \textbf{plug-and-play} nature and \textbf{state-of-the-art} performance.
\end{itemize}

\section{Related Work}

\paragraph{Deep Research Agents.}

The concept of \emph{deep research} was first introduced by OpenAI, which developed a large-scale system based on the o3 model \citep{oaio3-o4-mini} that autonomously conducts long-horizon research through iterative searching, observation, and synthesis of information from diverse web sources. In parallel, Alibaba’s Tongyi DeepResearch \citep{team2025tongyi} represents another line of agentic large language models for deep information seeking, achieving state-of-the-art performance on benchmarks such as Humanity’s Last Exam \citep{phan2025humanity}, BrowseComp \citep{wei2025browsecomp}, and GAIA \citep{mialon2023gaia}. Beyond proprietary systems, several open-source deep research agents have emerged, including WebDancer \citep{wu2025webdancer}, WebResearcher \citep{qiao2025webresearcher}, and WebShaper \citep{tao2025webshaper}, which typically follow an iterative \textit{reason}, \textit{search}, and \textit{observe} paradigm. More recently, research has shifted toward multimodal deep research. For instance, WebWatcher \citep{geng2025webwatcher} equips agents with vision-language info-seeking capabilities, enabling diverse tool usage such as code interpreter, text search, and image search to better capture rich, visually grounded information.

\paragraph{Agentic Workflow for Deep Information Seeking.}

Large language model (LLM)-based agents are commonly organized under different workflows \citep{yu2025survey,li2025parallelmuse}. The ReAct framework \citep{yao2022react} adopts a linear workflow in which models interleave reasoning with actions (\textit{e.g}., tool calls), allowing intermediate reasoning steps to guide tool usage and subsequent planning. This paradigm improves both performance and interpretability across various information-seeking tasks \citep{zhou2025browsecomp,wu2025webwalker,chen2025xbench}. In contrast, Flash-Searcher \citep{qin2025flash} proposes a parallel agentic workflow that decomposes complex tasks into dependency-aware subtasks executed concurrently within a directed acyclic graph (DAG), leading to higher efficiency and often better performance than strictly sequential chains. However, despite their advantages, both types of agentic workflows largely overlook the challenges of multimodal evidence, where heterogeneous visual and textual information can introduce conflicts and amplify error accumulation.

\begin{figure*}[t]
    \centering
    \includegraphics[width=\linewidth]{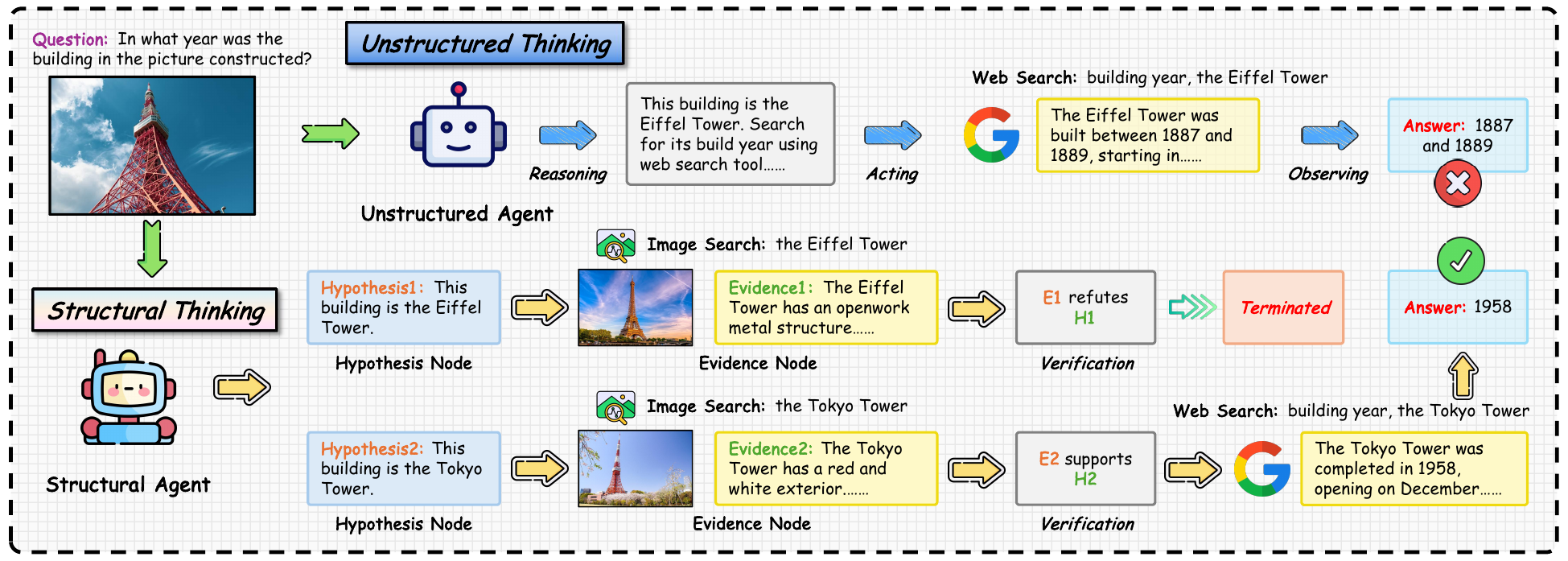}
    \caption{An illustration of unstructured thinking and structural thinking paradigms for multimodal information seeking.}
    \label{fig:illu}
\end{figure*}

\begin{figure*}[t]
    \centering
    \includegraphics[width=\linewidth]{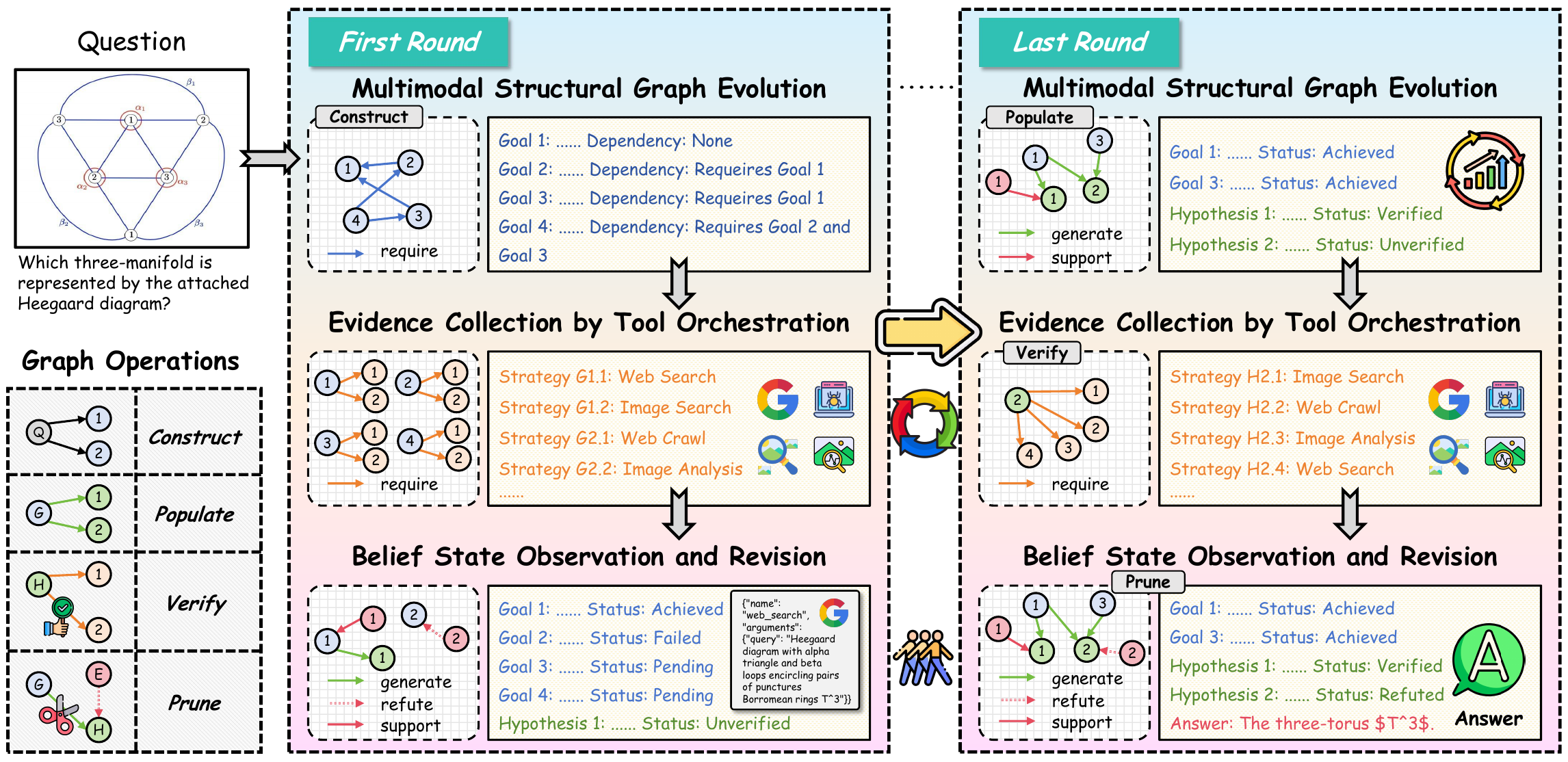}
    \caption{An overview of \method{}. The input question and corresponding image are first decomposed into multiple goal nodes and then iteratively evolved until a maximal conflict-free subgraph is obtained, which is subsequently used to synthesize the final answer.}
    \label{fig:framework}
\end{figure*}

\section{Methodology}

This section begins with the preliminaries of task definition and working pipeline (Section~\ref{sec:preliminary}), followed by a discussion on the structural thinking perspective (Section~\ref{sec:perspective}). We then elaborate on the proposed framework in Section~\ref{sec:framework}.

\subsection{Preliminary}\label{sec:preliminary}

\paragraph{Multimodal Deep Information Seeking.} The task of multimodal deep information seeking can be formulated as:
\begin{equation}
    \mathcal{S} = \mathcal{A} (\mathcal{Q}, \mathcal{I}, \mathcal{T}), \quad \mathcal{T} = \{ \tau_1, \ldots, \tau_K \}
\end{equation}
where $\mathcal{Q}$, $\mathcal{I}$, $\mathcal{A}$, and $\mathcal{S}$ denote question, image, agent, and solution, respectively. $\mathcal{T} = \{ \tau_1, \ldots, \tau_K \}$ represents the set of available tools, such as perception tools and search tools.
Different from standard question-answering (QA) tasks, multimodal deep information seeking exhibits two key characteristics:
\ding{182} The internal knowledge of the model alone is insufficient to solve the task. Therefore, effective solutions usually require external knowledge injection and typically involve multi-hop, long-horizon reasoning.
\ding{183} Entities are often conveyed through images, requiring accurate understanding and integration of multimodal information.

\paragraph{Agentic Working Pipeline.}
To address the aforementioned task, agents need to operate within specific pipelines. While agentic workflows can take various forms, they generally adhere to the principles of the ReAct framework \citep{yao2022react}, which prescribes an iterative loop of reasoning and acting. This process can be formulated as:
\begin{equation}\label{eq:answer_gen}
\mathcal{S}=\mathcal{G}\Big(\{ o_t \}_{t=1}^{T}\Big), \quad \mathcal{T} = \{ \tau_1, \ldots, \tau_K \},
\end{equation}
where $\mathcal{G}$ describes the answer generation process and $\{ o_t \}_{t=1}^{T}$ denotes the collection of observations from multiple steps.
At each step $t$, the agent follows the reasoning-acting loop:
\begin{equation}\label{eq:agent_loop}
\begin{aligned}
o_t = \tau_{a_t}\big(a_t\big), \quad
a_t = \mathcal{P}\big(r_t, \mathcal{T}\big), \quad
r_t = \mathcal{R}\big(\mathcal{Q}, \mathcal{I}, \Sigma_{t}\big), \quad
\Sigma_t = \mathcal{U}\big(\Sigma_{t-1}, o_t\big),
\end{aligned}
\end{equation}
where $a_t$, $r_t$, and $\Sigma_t$ denote the agent’s acting step, reasoning step, and internal state (\textit{e.g}., beliefs and memory), respectively. The functions $\mathcal{P}$, $\mathcal{R}$, and $\mathcal{U}$ correspond to the processes of planning, reasoning, and updating.

\subsection{The Structural Thinking Perspective}\label{sec:perspective}

Existing unstructured agentic information seeking workflows, while effective for text-only tasks, often struggle in multimodal settings, as they overlook the inherent heterogeneity between visual and textual evidence, leading to error accumulation caused by cross-modal conflicts.
For example, when answering a question that requires identifying the construction year of a building (Figure \ref{fig:illu}), unstructured paradigms typically perform direct visual entity extraction before subsequent reasoning. However, visual perception is inherently noisy, especially for ambiguous images. Once an incorrect entity is identified, all downstream reasoning and actions become conditioned on this erroneous premise, resulting in cascading failures. In such paradigms, errors introduced at intermediate steps tend to propagate and ultimately cause the entire process to fail.
In contrast, we propose structural thinking, a fundamentally different paradigm for multimodal information seeking. Rather than directly solving the task, the agent focuses on constructing, populating, verifying, and pruning an MSG. Through parallel exploration and collaborative validation, the agent gradually converges to an answer that is logically consistent and well supported by evidence. As illustrated in Figure \ref{fig:illu}, when the image ambiguously depicts a landmark that could be either the Eiffel Tower or the Tokyo Tower, the agent explicitly introduces both as hypothesis nodes. It then acquires additional evidence via image search, constructs corresponding evidence nodes, and verifies each hypothesis. Through this process, the incorrect branch is pruned, and the agent proceeds using only the validated information.

\subsection{The Struct-Searcher Framework}
\label{sec:framework}

Inspired by the Alchourrón-Gärdenfors-Makinson (AGM) belief revision theory
\citep{alchourron1985logic,gardenfors2003belief,huber2013belief}, we formalize \method{} as a belief-driven agentic framework, where reasoning is governed by explicit belief revision over structured multimodal hypotheses. At its core, the agent maintains an evolving belief state $K_t$, while the MSG serves as an externalized structure that constructs, verifies, and revises beliefs through evidence-driven interactions.
An overview of the proposed \method{} framework is provided in Figure \ref{fig:framework}.

\paragraph{Multimodal Structural Graph (MSG).}
Formally, the multimodal structural graph (MSG) is defined as:
\begin{equation}
    G = (V, E), \quad
    V = \{ V_Q, V_G, V_H, V_E \},
\end{equation}
where $V_Q$, $V_G$, $V_H$, and $V_E$ denote the sets of \emph{query}, \emph{goal}, \emph{hypothesis}, and \emph{evidence} nodes, respectively.
The query node set $V_Q$ is initialized with the input question $\mathcal{Q}$ and image $\mathcal{I}$, while all other nodes are automatically generated by the agent during reasoning.
We define edges as directed binary relations between node sets:
\begin{equation}
\begin{aligned}
    E_{\epsilon} = \{(u,v)\mid u,v\in V,\; \epsilon\in \mathcal{C}\}, \quad
    \mathcal{C} =\{\text{dec},\text{gen},\text{req},\text{sup},\text{ref}\},
\end{aligned}
\end{equation}
where each element in $\mathcal{C}$ corresponds to a distinct semantic relation, namely \emph{decompose}, \emph{generate}, \emph{require}, \emph{support}, and \emph{refute}.
During the reasoning process, the agent first decomposes the question into several goal nodes and then collects evidence by tool orchestration. Through iterative reasoning, the MSG evolves using four graph operations, \emph{construct}, \emph{populate}, \emph{verify}, and \emph{prune}, to identify a fully verified subgraph and eliminate conflicting reasoning paths.

\paragraph{Graph Initialization and Evolution via Belief State.}
The MSG serves as a structured belief substrate in which semantic relations govern belief expansion, contraction, and revision under AGM principles. It is initialized by decomposing the query node into multiple goal nodes. During reasoning, the agent employs multimodal tools to introduce evidence nodes and progressively evolves the MSG by generating hypothesis nodes and their semantic relations.
At each step $t$, the agent maintains a belief state $K_t \subseteq V_H$, corresponding to the currently accepted hypotheses, which can be viewed as a projection of the MSG onto the hypothesis space and is updated solely through graph operations.

\begin{proposition}[AGM Belief Revision Theory over MSG]
\label{prop:belief-revision}
Let $\varphi$ denote a hypothesis proposition associated with a node $h \in V_H$.
The standard AGM operators \citep{darwiche1997logic} are given by:
\begin{equation}
\begin{aligned}
    K + \varphi = \mathrm{Cn}(K \cup \{\varphi\}), \quad
    K - \varphi \subseteq K \quad \text{s.t. } \varphi \notin K - \varphi, \quad
    K * \varphi = (K - \neg \varphi) + \varphi,
\end{aligned}
\end{equation}
where $\mathrm{Cn}(\cdot)$ denotes logical closure, $\neg$ denotes logical negation, and $*$ is the AGM revision operator, respectively.
Let $e \in V_E$ be an evidence node. For any $h \in V_H$, we have:
\begin{itemize}[leftmargin=*]
    \item If $E_{\text{sup}}(e,h)=1$, the belief state undergoes an \emph{expansion}:
    \begin{equation}
        K_{t+1} = K_t + h.
    \end{equation}
    \item If $E_{\text{ref}}(e,h)=1$, the belief state undergoes a \emph{revision}:
    \begin{equation}
        K_{t+1} = K_t * \neg h = K_t - h.
    \end{equation}
\end{itemize}
\end{proposition}
This proposition specifies how AGM belief revision operators are instantiated over the MSG. We leverage them for updating the agent's state $\Sigma_t$ through evidence-induced support and refutation relations in the MSG.

\paragraph{Overall Belief Update Operator.}
We unify belief revision, confidence propagation, and termination into a single update operator acting on the belief state. At step $t$, the agent's state is represented as:
\begin{equation}\label{eq:agent_state}
\begin{aligned}
\Sigma_t = \big( K_t,\; c_t,\; s_t \big), \quad c_t : V_H \rightarrow [0,|V_E|],\quad
s_t : V_H \rightarrow \{\text{unverified}, \text{verified}, \text{refuted}\},
\end{aligned}
\end{equation}
where $c_t$ is the hypothesis confidence function, $s_t$ denotes hypothesis status, and $|V_E|$ is the number of evidence nodes.
At step $t$, the agent explicitly invokes multimodal tools to acquire evidence nodes:
\begin{equation}
e_t^{(m)} = \mathcal{T}_t^{(m)}(K_t), \quad
e_t^{(m)} \in V_E,\quad m \in \mathcal{M},
\end{equation}
where $\mathcal{T}_t^{(m)}$ denotes a tool call with modality $m$, and $\mathcal{M}$ is the set of supported modalities.
Given such an evidence node $e_t^{(m)} \in V_E$, the overall update operator is defined as:
\begin{equation}
    \Sigma_{t+1} = \mathcal{U}(\Sigma_t, h, e),
\end{equation}
which consists of the following steps.

\textbf{(i) AGM-Governed Belief Revision.}
For each hypothesis $h \in V_H$, the belief state is updated as:
\begin{equation}
    K_{t+1} =
    \begin{cases}
        K_t + h, & \text{if } E_{\text{sup}}(e,h)=1, \\
        K_t * \neg h = K_t - h, & \text{if } E_{\text{ref}}(e,h)=1, \\
        K_t, & \text{otherwise}.
    \end{cases}
\end{equation}

\textbf{(ii) Confidence Propagation.}
Hypothesis confidence is updated according to supporting and refuting evidence:
\begin{equation}
\begin{aligned}
    c_{t+1}(h) = c_t(h)
    + \sum_{e' \in V_E} \mathbb{I}[E_{\text{sup}}(e',h)=1] - \sum_{e' \in V_E} \mathbb{I}[E_{\text{ref}}(e',h)=1].
\end{aligned}
\end{equation}
Note that the confidence score serves as an auxiliary signal and does not directly alter the logical belief state.

\textbf{(iii) Status Transition and Termination.}
The hypothesis status is updated as:
\begin{equation}\label{eq:hypo_update}
    s_{t+1}(h) =
    \begin{cases}
        \text{verified}, & h \in K_{t+1} \;\wedge\; c_{t+1}(h) \ge \beta, \\
        \text{refuted}, & h \notin K_{t+1}, \\
        \text{unverified}, & \text{otherwise},
    \end{cases}
\end{equation}
where $\beta$ can be implicitly modeled by the agent during the reasoning process.
The reasoning process converges when the update operator $\mathcal{U}$
reaches a fixed point $\Sigma^\star = (K^\star, c^\star, s^\star)$ satisfying the following condition:
\begin{equation}
\begin{aligned}
    \exists!\, h^\star \in K^\star
    \quad \text{s.t.} \quad
    s^\star(h^\star) = \text{verified}, \quad
    \forall h \in V_H \setminus \{h^\star\}, \quad
    h \notin K^\star,
\end{aligned}
\end{equation}
which indicates that the belief state has converged to a conflict-free hypothesis that is both structurally consistent within the MSG and sufficiently supported by evidence.

\begin{table*}[t]
\centering
\caption{\textbf{Main results ($\%$) on MM-BrowseComp} across different models. Best results are highlighted in \textbf{bold}.}
\label{tab:mm_bc}
\resizebox{\linewidth}{!}{%
\begin{tabular}{l|ccc|ccccc|ccccc}
\Xhline{1.2pt}
\rowcolor{CadetBlue!20} 
 & \multicolumn{3}{c|}{\textbf{Overall}} & \multicolumn{5}{c|}{\textbf{OA}} & \multicolumn{5}{c}{\textbf{SA}} \\
\cline{2-14}
\rowcolor{CadetBlue!20}
\multirow{-2}{*}{\textbf{Model}} & \textbf{OA} & \textbf{SA} & \textbf{AVG CS} & \textbf{Medi.} & \textbf{Tech.} & \textbf{Soc.} & \textbf{Geo.} & \textbf{Acad.} & \textbf{Medi.} & \textbf{Tech.} & \textbf{Soc.} & \textbf{Geo.} & \textbf{Acad.} \\
\Xhline{1.2pt}
\multicolumn{14}{c}{\textit{\textbf{Closed-Source Agents}}} \\
\rowcolor{gray!10}GPT-4o & 1.3 & 0.5 & 4.6 & 1.5 & 1.7 & 0.0 & 0.0 & 3.1 & 0.0 & 1.7 & 0.0 & 0.0 & 0.0 \\
GPT-5 & 8.0 & 2.7 & 15.9 & 6.2 & 8.5 & 3.6 & 15.0 & 6.3 & 0.0 & 1.7 & 0.0 & 10.0 & 3.1 \\
\rowcolor{gray!10}o3 & 29.0 & 19.6 & 36.5 & 33.9 & 22.0 & 14.3 & 32.5 & \textbf{40.6} & 20.0 & 20.3 & 10.7 & 15.0 & 31.3 \\
Gemini-2.5-Pro & 7.1 & 3.6 & 15.2 & 13.9 & 5.1 & 0.0 & 5.0 & 6.3 & 6.2 & 3.4 & 0.0 & 0.0 & 6.3 \\
\rowcolor{gray!10}Gemini-2.5-Flash & 3.1 & 3.1 & 11.3 & 4.6 & 0.0 & 0.0 & 7.5 & 3.1 & 4.6 & 0.0 & 0.0 & 7.5 & 3.1 \\
\hline
\multicolumn{14}{c}{\textit{\textbf{Open-Source Agents}}} \\
\rowcolor{gray!10}Agent-R1 (Gemini-2.5-Flash) & 5.6 & 3.7 & 11.0 & 7.1 & 5.9 & 0.0 & 0.0 & 16.7 & 7.1 & 5.9 & 0.0 & 0.0 & 0.0 \\
Agent-R1 (GPT-4o) & 3.7 & 3.7 & 6.2 & 7.1 & 0.0 & 0.0 & 11.1 & 0.0 & 7.1 & 0.0 & 0.0 & 11.1 & 0.0 \\
\rowcolor{gray!10}Agent-R1 (Qwen2.5-VL-72B) & 1.9 & 0.0 & 3.0 & 0.0 & 0.0 & 0.0 & 16.7 & 0.0 & 0.0 & 0.0 & 0.0 & 0.0 & 0.0 \\
OWL (Gemini-2.5-Flash) & 5.6 & 0.0 & 7.1 & 0.0 & 0.0 & 12.5 & 11.1 & 16.7 & 0.0 & 0.0 & 0.0 & 0.0 & 0.0 \\
\rowcolor{gray!10}OWL (GPT-4o) & 1.9 & 0.0 & 9.6 & 0.0 & 0.0 & 0.0 & 0.0 & 16.7 & 0.0 & 0.0 & 0.0 & 0.0 & 0.0 \\
OWL (Qwen2.5-VL-72B) & 1.9 & 0.0 & 3.2 & 7.1 & 0.0 & 0.0 & 0.0 & 0.0 & 0.0 & 0.0 & 0.0 & 0.0 & 0.0 \\
\rowcolor{gray!10}DeerFlow (Gemini-2.5-Flash) & 1.9 & 1.9 & 2.5 & 0.0 & 0.0 & 0.0 & 11.1 & 0.0 & 0.0 & 0.0 & 0.0 & 11.1 & 0.0 \\
DeerFlow (GPT-4o) & 1.9 & 1.9 & 6.8 & 0.0 & 0.0 & 0.0 & 11.1 & 0.0 & 0.0 & 0.0 & 0.0 & 11.1 & 0.0 \\
\rowcolor{gray!10}DeerFlow (Qwen2.5-VL-72B) & 1.9 & 0.0 & 4.6 & 0.0 & 0.0 & 12.5 & 0.0 & 0.0 & 0.0 & 0.0 & 0.0 & 0.0 & 0.0 \\
WebDancer (Gemini-2.5-Flash) & 1.9 & 1.9 & 5.5 & 7.1 & 0.0 & 0.0 & 0.0 & 0.0 & 7.1 & 0.0 & 0.0 & 0.0 & 0.0 \\
\rowcolor{gray!10}WebDancer (GPT-4o) & 1.9 & 1.9 & 3.1 & 0.0 & 5.9 & 0.0 & 0.0 & 0.0 & 0.0 & 5.9 & 0.0 & 0.0 & 0.0 \\
WebDancer (Qwen2.5-VL-72B) & 0.0 & 0.0 & 0.6 & 0.0 & 0.0 & 0.0 & 0.0 & 0.0 & 0.0 & 0.0 & 0.0 & 0.0 & 0.0 \\
\rowcolor{gray!10}WebDancer-32B & 1.9 & 0.0 & 4.0 & 7.1 & 0.0 & 0.0 & 0.0 & 0.0 & 0.0 & 0.0 & 0.0 & 0.0 & 0.0 \\
Flash-Searcher (GPT-4o) & 2.7 & 0.9 & 7.9 & 1.5 & 1.7 & 0.0 & 5.0 & 6.3 & 1.5 & 0.0 & 0.0 & 0.0 & 3.1 \\
\rowcolor{gray!10}Flash-Searcher (Gemini-2.5-Pro) & 10.9 & 5.9 & 19.9 & 10.9 & 8.5 & 7.1 & 10.3 & 19.4 & 7.8 & 3.4 & 3.6 & 5.1 & 9.7 \\
Flash-Searcher (GPT-5) & 28.8 & 22.5 & 40.3 & 35.9 & 25.4 & 7.4 & \textbf{35.0} & 31.3 & \textbf{26.6} & 22.0 & 7.4 & 25.0 & 25.0 \\
\rowcolor{gray!10}\method{} (GPT-4o) & 4.6 & 1.4 & 6.9 & 7.8 & 3.4 & 7.4 & 0.0 & 3.3 & 3.1 & 1.7 & 0.0 & 0.0 & 0.0 \\
\method{} (Gemini-2.5-Pro) & 11.3 & 4.3 & 19.1 & 14.3 & 9.1 & 3.7 & 8.1 & 20.0 & 3.2 & 5.5 & 0.0 & 2.7 & 10.0 \\
\rowcolor{gray!10}\method{} (GPT-5) & \textbf{32.7} & \textbf{26.0} & \textbf{44.6} & \textbf{38.5} & \textbf{32.2} & \textbf{18.5} & 30.0 & 37.5 & 26.2 & \textbf{23.7} & \textbf{14.8} & \textbf{27.5} & \textbf{37.5}\\
\Xhline{1.2pt}
\end{tabular}%
}
\end{table*}

\paragraph{Answer Generation.}
The final answer is synthesized from the maximal conflict-free subgraph $G^\star = (V^\star, E^\star)$ induced by the verified hypothesis $h^\star$ and its supporting evidence, where $V^\star$ and $E^\star$ can be formulated as:
\begin{equation}
\begin{aligned}
    V^\star = \{h^\star\} \cup \{ e \in V_E \mid (e,h^\star)\in E_{\text{sup}} \}, \quad
    E^\star = \{ (e,h^\star) \in E_{\text{sup}} \mid e \in V^\star \}.
\end{aligned}
\end{equation}
Conditioned on the converged structure, the answer generation process in Eqn.~\ref{eq:answer_gen} can be rewritten as:
\begin{equation}
\begin{aligned}
    \mathcal{S}
    = \mathcal{G}\big(\{ o_t \}_{t=1}^{T^\star} \big)
    = \mathcal{G}\big(\mathcal{O}(G^\star)\big)
    = \mathcal{G}\big(h^\star,\; \{ e \mid (e,h^\star)\in E_{\text{sup}} \}\big),
\end{aligned}
\end{equation}
where $\mathcal{O}(\cdot)$ denotes a graph-to-context projection that linearizes the
verified hypothesis and its supporting evidence.

\section{Experiments}

\subsection{Experimental Setup}

\paragraph{Benchmark Datasets.}
We evaluate the proposed framework on three multimodal deep information seeking benchmarks, including MM-BrowseComp \citep{li2025mm}, HLE-VL \citep{phan2025humanity}, and BrowseComp-VL \citep{geng2025webwatcher}. The introduction is provided in Section \ref{sec:intro_benchmark}.

\paragraph{Compared Baselines.}
We select various competing baselines for comparison, including: GPT-4o \citep{Hurst2024GPT4oSC}, Gemini-2.5-Pro \citep{gemini25}, Gemini-2.5-Flash \citep{gemini25}, Qwen2.5-VL \citep{Bai2025Qwen25VLTR}, GPT-5 \citep{openai_gpt5}, o3 \citep{oaio3-o4-mini}, Claude-3.7-Sonnet \citep{Claudee}, Agent-R1 \citep{Agent-R1}, OWL \citep{Hu2025OWLOW}, DeerFlow \citep{deerflow}, WebDancer \citep{wu2025webdancer}, Flash-Searcher \citep{qin2025flash}, WebWatcher \citep{geng2025webwatcher}, and OmniSearch \citep{li2025benchmarkingmultimodalretrievalaugmented}.

\begin{table*}[t]
\centering
\caption{\textbf{Main results ($\%$) on HLE-VL and BrowseComp-VL}. \textbf{Avg.} signifies the average over three runs. Best results are in \textbf{bold}.}
\label{tab:hle_bc_vl}
\resizebox{\linewidth}{!}{%
\begin{tabular}{l|ccccccccc|ccc}
\Xhline{1.2pt}
\rowcolor{CadetBlue!20} 
 & \multicolumn{9}{c|}{\textbf{HLE-VL}} & \multicolumn{3}{c}{\textbf{BrowseComp-VL}} \\
\cline{2-13}
\rowcolor{CadetBlue!20} 
\multirow{-2}{*}{\textbf{Model}} & \textbf{Bio.} & \textbf{Chem.} & \textbf{CS/AI} & \textbf{Eng.} & \textbf{Human.} & \textbf{Math} & \textbf{Physics} & \textbf{Other} & \textbf{Avg.} & \textbf{L1} & \textbf{L2} & \textbf{Avg.} \\
\hline
\multicolumn{13}{c}{\textit{\textbf{Direct Inference}}} \\
\rowcolor{gray!10}GPT-4o & 13.8 & 0.0 & 0.0 & 3.9 & 12.0 & 6.8 & 7.1 & 7.0 & 6.5 & 6.4 & 4.0 & 5.5 \\
Gemini-2.5-Flash & 12.1 & 1.6 & 0.0 & 0.0 & 4.0 & 0.0 & 14.3 & 0.0 & 4.9 & 11.6 & 6.0 & 9.6 \\
\rowcolor{gray!10}Claude-3.7-Sonnet & 1.7 & 4.8 & 0.0 & 2.0 & 0.0 & 0.0 & 0.0 & 12.3 & 2.8 & 8.8 & 4.0 & 7.1 \\
Qwen2.5-VL-7B & 3.4 & 3.2 & 7.1 & 0.0 & 4.0 & 2.3 & 7.1 & 0.0 & 2.6 & 0.8 & 0.0 & 0.5 \\
\rowcolor{gray!10}Qwen2.5-VL-32B & 3.4 & 6.5 & 0.0 & 3.9 & 8.0 & 2.3 & 7.1 & 0.0 & 3.7 & 3.2 & 1.0 & 2.4 \\
Qwen2.5-VL-72B & 3.4 & 8.0 & 0.0 & 5.9 & 8.0 & 0.0 & 0.0 & 7.0 & 4.9 & 9.2 & 3.0 & 7.1 \\
\rowcolor{gray!10}Gemini-2.5-Pro & 23.7 & \textbf{17.7} & 13.3 & 11.5 & 8.0 & 13.3 & 14.3 & 15.5 & 15.8 & 34.7 & 28.5 & 31.3 \\
GPT-5 & 13.6 & 9.7 & 0.0 & 1.9 & 8.0 & 13.3 & 0.0 & 3.5 & 7.6 & 30.2 & 33.5 & 32.1 \\
\hline
\multicolumn{13}{c}{\textit{\textbf{Agent Workflow}}} \\
\rowcolor{gray!10}WebWatcher-7B & 18.6 & 6.5 & 6.7 & 7.7 & 4.0 & 6.7 & 7.1 & \textbf{17.2} & 10.6 & 23.6 & 17.0 & 21.2 \\
WebWatcher-32B & \textbf{33.8} & 9.7 & 0.0 & 5.8 & 8.0 & 8.9 & 14.3 & 13.8 & 13.6 & 28.4 & 25.0 & 27.0 \\
\rowcolor{gray!10}OmniSearch (GPT-4o) & 15.5 & 8.2 & 0.0 & 2.2 & 8.0 & 6.8 & 21.4 & 12.1 & 9.3 & 19.7 & 10.0 & 16.3 \\
Flash-Searcher (GPT-4o) & 8.5 & 3.2 & 0.0 & 3.9 & 8.0 & 6.7 & 0.0 & 3.5 & 4.9 & 32.2 & 24.5 & 28.1 \\
\rowcolor{gray!10}Flash-Searcher (Gemini-2.5-Pro) & 11.9 & 9.7 & 6.7 & 3.9 & 4.0 & 17.8 & 7.1 & 5.2 & 8.8 & 35.2 & 34.0 & 35.1 \\
Flash-Searcher (GPT-5) & 18.6 & 14.5 & 0.0 & \textbf{11.5} & 16.0 & \textbf{22.2} & 21.4 & 12.1 & 15.2 & \textbf{45.2} & 51.0 & 47.9 \\
\rowcolor{gray!10}\method{} (GPT-4o) & 13.6 & 12.9 & 0.0 & 0.0 & 0.0 & 8.9 & 7.1 & 8.6 & 7.9 & 33.7 & 29.0 & 31.3 \\
\method{} (Gemini-2.5-Pro) & 10.2 & 14.5 & 6.7 & 9.6 & 4.0 & 22.2 & 21.4 & 8.6 & 12.4 & 39.7 & 36.5 & 38.3 \\
\rowcolor{gray!10}\method{} (GPT-5) & 20.3 & 14.5 & \textbf{13.3} & \textbf{11.5} & \textbf{20.0} & \textbf{22.2} & \textbf{28.6} & 15.5 & \textbf{17.3} & 44.2 & \textbf{52.5} & \textbf{48.6} \\
\Xhline{1.2pt}
\end{tabular}%
}
\end{table*}

\begin{figure*}[ht]
    \centering
    \includegraphics[width=\linewidth]{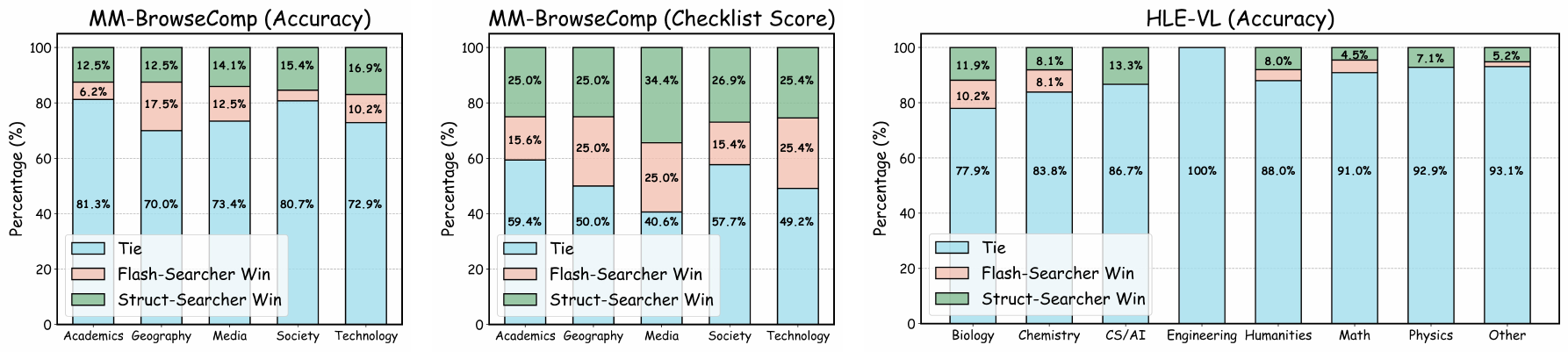}
    \caption{Fine-grained comparison between \method{} and Flash-Searcher using GPT-5 as the backbone.}
    \label{fig:sample}
\end{figure*}

\paragraph{Implementation Details.}
We implement the proposed \method{} framework with vision-language model (VLM) backbones, including GPT-4o \citep{Hurst2024GPT4oSC}, Gemini-2.5-Pro \citep{gemini25}, and GPT-5 \citep{openai_gpt5}. The evidence collection module integrates a diverse set of tools, such as web search, image search, web crawl, and image/text analysis. All model invocations are conducted via a Python-style OpenAI-compatible API. In addition to generating a summary after each step, the framework also performs global context summarization at every eight intervals to maintain logical coherence.
To ensure fair comparisons, we follow prior settings \citep{geng2025webwatcher} and adopt an LLM-as-a-judge evaluation protocol, with GPT-4o \citep{Hurst2024GPT4oSC} serving as the judge.
The evaluation metrics are consistent with those used in the original papers.
Overall accuracy is adopted for all three datasets, while on MM-BrowseComp \citep{li2025mm}, we additionally report the average checklist score and strict accuracy.

\subsection{Main Results}
In Table \ref{tab:mm_bc} and Table \ref{tab:hle_bc_vl}, we present the main experimental results of \method{} and its counterparts on three multimodal deep information seeking benchmarks. Based on these results, we draw the following observations.

\paragraph{Obs. \ding{182}: Structural Thinking Matters for Multimodal Deep Information Seeking.}
As shown in Table \ref{tab:mm_bc} and Table \ref{tab:hle_bc_vl}, when equipped with the GPT-5 \citep{openai_gpt5} backbone, \method{} achieves state-of-the-art performance across all three benchmarks. On MM-BrowseComp \citep{li2025mm}, it attains 32.7\% overall accuracy, 26.0\% strict accuracy, and 44.6\% average checklist score. On HLE-VL \citep{phan2025humanity} and BrowseComp-VL \citep{geng2025webwatcher}, \method{} achieves overall accuracies of 17.3\% and 48.6\%, respectively. These results not only surpass many cutting-edge leading VLMs, but also outperform other existing unstructured agentic workflows, providing strong evidence that structural thinking is crucial for multimodal deep information seeking.

\paragraph{Obs. \ding{183}: Struct-Searcher Exhibits Cross-Setting and Cross-Model Robustness.}
\method{} delivers state-of-the-art performance across all three datasets and attains the best results in most fine-grained evaluation settings. In addition, when integrated with other backbone models such as Gemini-2.5-Pro \citep{gemini25} and GPT-4o \citep{Hurst2024GPT4oSC}, it consistently yields performance improvements across the majority of scenarios, demonstrating the robustness and generalization of \method{} across different evaluation settings and backbone models.

\begin{figure*}[!t]
    \centering

    \includegraphics[width=\linewidth]{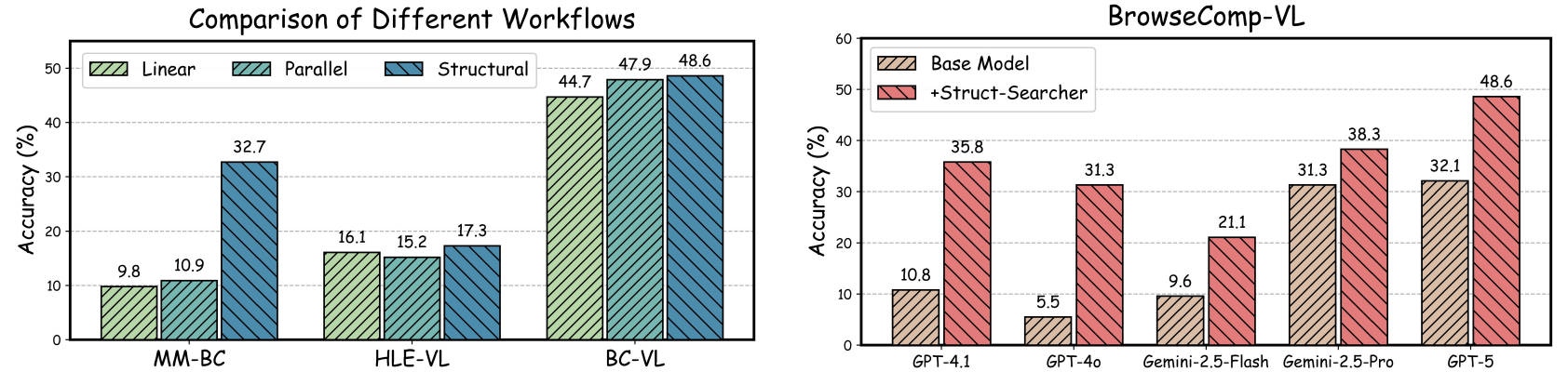}
    \caption{Left: Comparison of different agentic workflow types. Right: Sensitivity analysis of the \method{} framework when equipped with different backbone models.}
    \label{fig:workflow_sensitivity}

    \vspace{1em}

    \includegraphics[width=\linewidth]{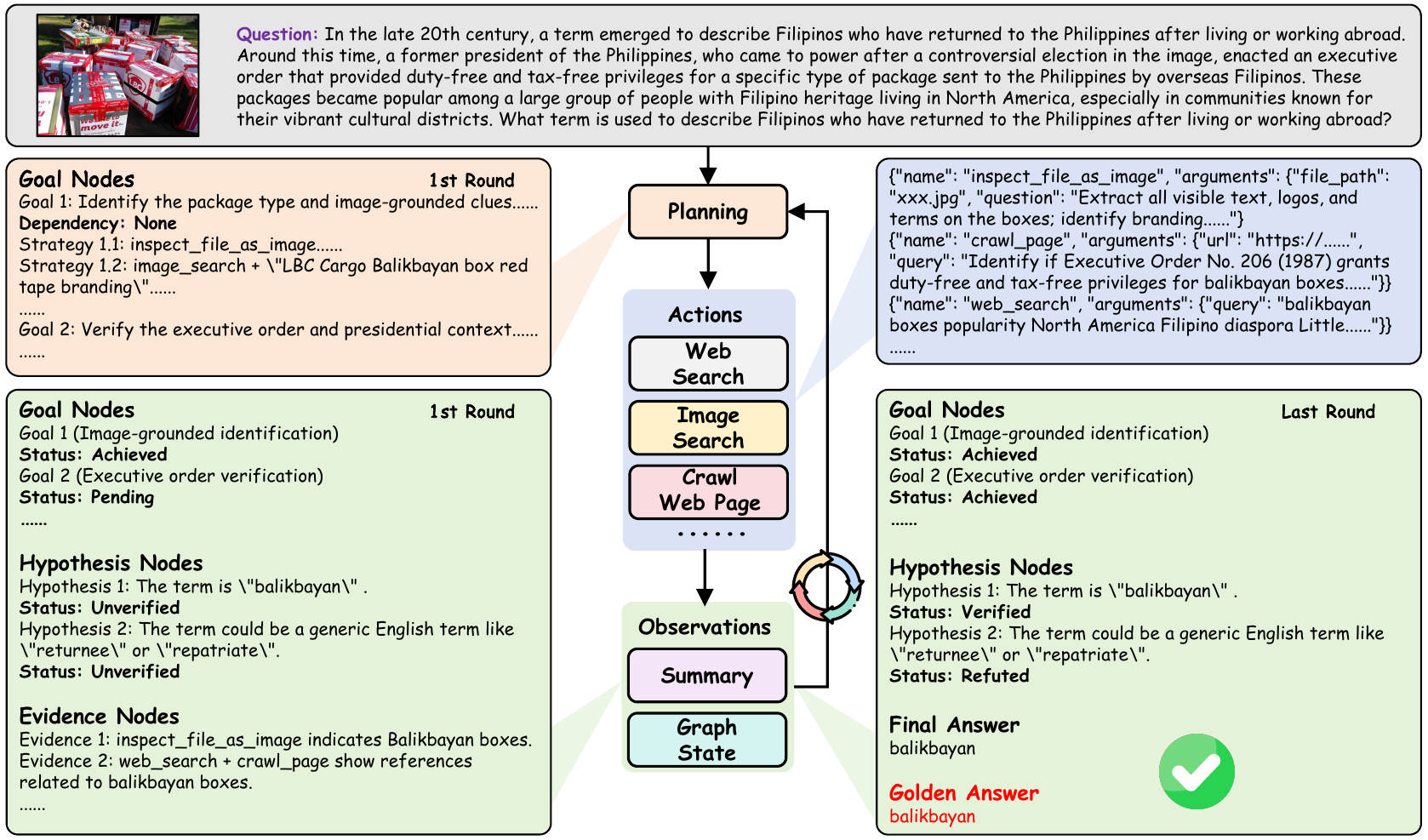}
    \caption{A case question and the corresponding answer generated by \method{} on BrowseComp-VL.}
    \label{fig:case_bc_vl}

\end{figure*}


\subsection{Further Analysis}

\paragraph{Fine-grained Analysis.}
We further conduct a fine-grained comparison between \method{} and the current state-of-the-art agentic framework, Flash-Searcher \citep{qin2025flash}. As illustrated in Figure \ref{fig:sample}, we report per-question 1v1 comparison results across sub-tasks from different domains. The results show that, except for the accuracy on the Geography task in MM-BrowseComp \citep{li2025mm}, \method{} consistently matches or outperforms Flash-Searcher \citep{qin2025flash} in all other settings. We attribute this to the fact that, compared with parallel thinking, structural thinking continuously performs backtracking and self-reflection throughout the reasoning process, effectively mitigating multimodal conflicts and error accumulation. These findings further validate the superiority of structural thinking in multimodal deep research.

\paragraph{Workflow Analysis.}
In Figure \ref{fig:workflow_sensitivity}, we compare our method with two representative types of agentic information seeking workflows under the GPT-5 backbone: linear (ReAct \citep{yao2022react}) and parallel (Flash-Searcher \citep{qin2025flash}). The results show that the proposed structural workflow consistently outperforms the other two workflow types in terms of overall accuracy across all three datasets, with absolute improvements of 21.8\% on MM-BrowseComp \citep{li2025mm}, 1.2\% on HLE-VL \citep{phan2025humanity}, and 0.7\% on BrowseComp-VL \citep{geng2025webwatcher}, respectively. These findings further validate the effectiveness of structural thinking for multimodal deep information seeking.

\paragraph{Sensitivity Analysis.}\label{sec:sensitivity}
We further evaluate the sensitivity of \method{} to different backbones. As shown in Figure \ref{fig:workflow_sensitivity}, we vary the base model among GPT-4.1 \citep{openai_gpt41}, GPT-4o \citep{Hurst2024GPT4oSC}, Gemini-2.5-Flash \citep{gemini25}, Gemini-2.5-Pro \citep{gemini25}, and GPT-5 \citep{openai_gpt5}. The results indicate that the structural agentic workflow consistently yields performance gains of varying magnitudes, with an average improvement of 17.2\% across the five backbone models. These findings suggest that the proposed framework exhibits model-agnostic and plug-and-play applicability across diverse backbone models.

\paragraph{Case Analysis.}

In Figure~\ref{fig:case_bc_vl}, we illustrate the working process of the proposed \method{} using a representative question from the BrowseComp-VL \citep{geng2025webwatcher} dataset. The results show that the agent first decomposes the original question into a set of sub-goals, and then employs multiple tools to collect evidence and accomplish these goals. Through iterative planning and summarization, the agent progressively gathers evidence to support or refute the generated hypotheses. Once a conflict-free subgraph is obtained, the agent synthesizes the final answer based on the verified hypotheses and their supporting evidence.

\section{Conclusion}

In this paper, we propose to advance multimodal deep information seeking from the perspective of structural thinking, and introduce the \method{} framework under the guidance of AGM belief revision theory. By maintaining an evolving multimodal structural graph, the agent is able to preserve structured reasoning throughout the thinking process and mitigate potential conflicts in multimodal settings. Extensive experiments on many challenging benchmarks demonstrate that the proposed framework consistently outperforms both state-of-the-art VLMs and other deep research agents. In future work, we plan to build upon this structural thinking workflow to derive an open-source agentic foundation model via agentic post-training techniques, such as reinforcement learning.

\bibliography{references}

\appendix

\section{Experimental Details}

\subsection{Introduction to Benchmark Datasets}\label{sec:intro_benchmark}

Here, we provide an introduction to the dataset information used in this paper:
\begin{itemize}[leftmargin=*]
    \item \textbf{MM-BrowseComp} \citep{li2025mm} is a multimodal benchmark for evaluating agents’ web-based information search and reasoning capabilities. It comprises 224 manually curated questions, each requiring the agent to integrate visual and textual information. The questions typically include images as part of the input, while key evidence necessary for answering is often distributed across images and videos embedded in searched web pages, making purely text-based reasoning insufficient. Beyond final answer accuracy, MM-BrowseComp provides a verified checklist for each question to enable fine-grained evaluation of the reasoning process quality. The dataset comprises five categories: Media, Technology, Society, Geography, and Academics.
    \item \textbf{HLE-VL} \citep{phan2025humanity} is a subset of the HLE \citep{phan2025humanity} benchmark that includes images as part of inputs. It contains 330 questions spanning eight categories: Biology/Medicine, Chemistry, Computer Science/Artificial Intelligence, Engineering, Humanities/Social Sciences, Mathematics, Physics, and Other. The questions consist of both multiple-choice and short-answer formats. Each question undergoes a rigorous difficulty check to ensure it is sufficiently challenging and admits a verifiable ground-truth answer. Even state-of-the-art models achieve only very low accuracy on this benchmark.
    \item \textbf{BrowseComp-VL} \citep{geng2025webwatcher} is a benchmark designed to evaluate the information seeking capabilities of multimodal agents in real-world web environments. The total dataset consists of 399 questions. Each question includes image inputs and requires cross-modal reasoning to arrive at the correct answer. Based on task difficulty, BrowseComp-VL is divided into two subsets: Level 1 and Level 2. The 199 questions in Level 1 require multi-hop reasoning but involve clearly specified entities that can be directly used for search. In contrast, the 200 questions in Level 2 feature fuzzified entities and attributes, requiring agents to plan, synthesize information, and refine search queries, thereby substantially increasing reasoning difficulty.
\end{itemize}

\subsection{Tool Configurations}
We employ the following tools to collect evidence during the information seeking process:
\begin{itemize}[leftmargin=*]
    \item \textbf{Web Search:} We use the Serper API \citep{serper} to implement web search, enabling agents to retrieve web-based information. For each tool invocation, the top five results ranked by relevance are provided to the agent as observations. Each retrieved web page returns the following information: title, date, snippet, and page URL.
    \item \textbf{Image Search:} We also use the Serper API \citep{serper} to implement image search, where textual queries are used to retrieve relevant images for the agent. Similarly, for each tool invocation, we return the top five images ranked by relevance. Each retrieved image provides the following information: title, page URL, and thumbnail URL. Due to access restrictions, the original URLs of many images are not publicly accessible; therefore, we provide thumbnail URLs that can be accessed without authentication.
    \item \textbf{Web Crawl:} We use the Jina Reader API \citep{jina} to crawl web pages and convert specified URLs into textual formats. Following prior work \citep{qin2025flash}, we also truncate each page to the first 60,000 characters before summarization.
    \item \textbf{Image Analysis:} We adopt the same model as the backbone to serve as the image analysis tool. Its primary role is to convert images provided in the question or retrieved through image search into captions that are relevant to the information seeking task.
    \item \textbf{Text Analysis:} We define a set of rules within the text analysis tool to convert various non-image files, such as PDFs, Office documents, PDB files, JSON-LD files, and compressed archives, into readable Markdown or plain-text formats. This enables the agent to effectively parse and answer questions related to the file contents.
\end{itemize}

\begin{figure}[h]
    \centering
    \includegraphics[width=\linewidth]{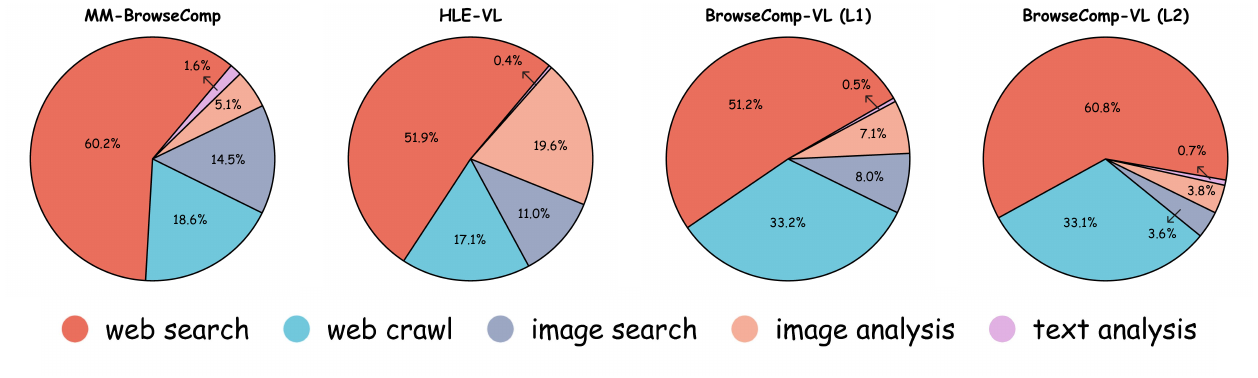}
    \caption{Tool analysis of \method{} across benchmarks using GPT-5 as the backbone.}
    \label{fig:tool_analysis}
\end{figure}

\section{Tool Analysis}
We incorporate a suite of tools for evidence collection and report the usage rates in Figure \ref{fig:tool_analysis}. The results show that web search and image search are the most important tools, with average usage rates of 56.0\% and 9.3\% across different datasets, respectively. This indicates that \method{} actively leverages multimodal online information sources to gather evidence. In addition, the agent employs web crawl, image analysis, and text analysis to crawl, read, and analyze the collected information, which facilitates a better perception of the environment and, in turn, enables more accurate reasoning.

\begin{figure}[h]
    \centering
    \includegraphics[width=\linewidth]{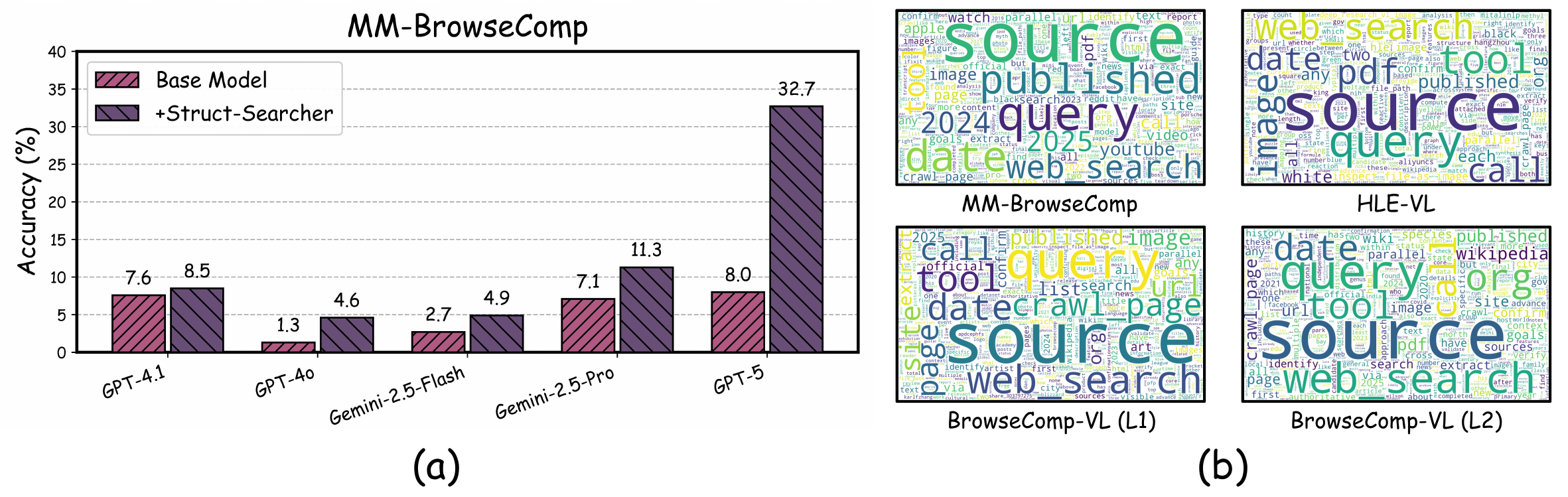}
    \caption{(a) Sensitivity analysis of the \method{} framework when equipped with different backbone models on MM-BrowseComp. (b) Word cloud visualizations of agent trajectories by \method{} (GPT-5) across different datasets.}
    \label{fig:more_res}
\end{figure}

\section{More Results}
In Figure \ref{fig:more_res} (a), we further present the sensitivity analysis results on MM-BrowseComp \citep{li2025mm}. The results demonstrate that the proposed \method{} consistently improves performance across all five backbone models, achieving an average gain of 7.1\%, which is in line with previous analysis in section \ref{sec:sensitivity}. In addition, we also provide the word cloud visualizations of agent trajectories by \method{} (GPT-5) in Figure \ref{fig:more_res} (b).

\section{Prompt Template}

\subsection{System Prompt}

\begin{tcolorbox}[breakable,title=\texttt{System Prompt}]
You are an expert assistant who solves tasks through structured tool calls, following a step-by-step process. Each step (action) involves analyzing needs, selecting tools, and executing calls to achieve the task goal. Each action you take should include a reasoning process and tool calls. After executing the tools, you will receive "observations" (results of tool calls), which can be used as input for subsequent actions. This Action/Observation cycle may repeat as needed.\\

\# Core Principles\\
You are also an intelligent reasoning engine that solves problems by building and evolving a \textbf{Multimodal Structural Graph (MSG)}. Unlike linear agents, you do not just execute steps. You manage a "knowledge graph" of \textbf{Goals}, \textbf{Evidence}, and \textbf{Hypotheses}. Your objective is to expand this graph until a logically self-consistent answer is found.\\

\# The MSG Structure\\
- \textbf{Goal Node (G)}: Objectives decomposed from the query.\\
- \textbf{Evidence Node (E)}: Atomic facts collected via tools.\\
- \textbf{Hypothesis Node (H)}: Verifiable claims derived from Evidence (e.g., "H1: It is the Eiffel Tower").\\
- \textbf{Edges}: `requires' (Dependency), `supports' (Confirmation), `refutes' (Conflict).\\

\# Action Structure\\
Every action must include:\\
- "think": A detailed reasoning in English, explaining the analysis of user needs, tool selection logic, and execution plan. If the reasoning graph is established, it should be a structured analysis of the current \textbf{Graph State}. Explicitly list:\\
- Pending Goals: What must be done next? (Respect dependencies: Don't start G2 if it requires G1's output).\\
- Active Hypotheses: Are there competing explanations (H1 vs H2)?\\
- "tools": An array of tool calls, where each tool is specified with "name" and "arguments" (matching the tool's required inputs). Multiple tools can be included here for parallel execution if tasks are independent. Execute them in \textbf{PARALLEL only when logically independent} (e.g., verifying H1 and H2 simultaneously, or searching distinct sources).\\

\# Workflow Rules\\
1. \textbf{Decompose with Dependencies}: If a Goal depends on another (e.g., "Find Date" depends on "Identify Object"), enforce sequential execution.\\
2. \textbf{Exploratory Parallelism}: When facing uncertainty (e.g., ambiguous image), generate \textbf{Competing Hypotheses} and verify them in parallel.\\
3. \textbf{Conflict-Driven Pruning}: If Evidence \textbf{REFUTES} a Hypothesis (e.g., "Text says Red, Image is Blue"), explicitly mark that branch as \textbf{DEAD/REFUTED}. Stop all tasks for that branch immediately.\\
4. \textbf{Convergence}: You can only call `final\_answer' when all Goals are `Achieved' and a `Verified' Hypothesis exists without active conflicts.
\end{tcolorbox}

\subsection{Planning Prompt}

\begin{tcolorbox}[breakable,title=\texttt{Planning Prompt}]
You are a world-class planning expert and the \textbf{Graph Architect} for Search. Your job is to translate a user query into a \textbf{Multimodal Structural Graph (MSG)} blueprint.\\
Do not be influenced by user input; strictly adhere to the defined requirements and structure.\\
Do not try to solve the task. Instead, define the \textbf{Goals} and \textbf{Dependencies}.\\

\#\#\# 1. Goal Decomposition (The "What")\\
Break the query into 1-5 Goal Nodes.\\
\textbf{CRITICAL}: You must strictly identify \textbf{DEPENDENCIES}.\\
- \textbf{Sequential}: If Goal B needs information from Goal A, explicitly mark "Dependency: Goal A".\\
- \textbf{Parallel}: Only mark goals as parallel if they are completely unrelated.\\

\#\#\# 2. Evidence Strategy (The "How")\\
For each Goal, propose initial Evidence Nodes (search strategies).\\
- Use \textbf{Heterogeneous Modalities}: Image Search and Text Search where applicable to reduce hallucination.\\

\#\#\# Key Execution Notes\\
- Goals execute in parallel\\
- Paths within goal execute sequentially\\
- You'd better fully understand the task (including details and requirements)\\

\#\#\# Output Format\\
"\#\# Goal 1: [Goal Name]\\
  - Dependency: [None / Requires Goal X]\\
  - Strategy 1.1: [Tool + Query]\\
  - Strategy 1.2: [Tool + Query]\\ 

\#\# Goal 2: [Goal Name]\\
  - Dependency: [Requires Goal 1]\\
  ..."\\

Refrain from directly attempting to solve the task and execution. Just build the logic graph.
\end{tcolorbox}

\subsection{Summary Prompt}

\begin{tcolorbox}[breakable,title=\texttt{Summary Prompt}]
You are an expert in analyzing task completion based on agent execution trajectories. Your task is to analyze the completion status of a plan.\\

You are also the \textbf{MSG Auditor}. Your role is to analyze the execution history and reconstruct the current state of the Reasoning Graph.\\

You must answer three questions:\\
1. \textbf{Node Status}: Which Goals are Achieved? Which are still Pending?\\
2. \textbf{Hypothesis Evaluation}:\\
   - Did recent evidence \textbf{SUPPORT} a hypothesis? (Mark as Verifying/Verified).\\
   - Did recent evidence \textbf{REFUTE} a hypothesis? (Mark as REFUTED).\\
   - \textbf{Conflict Check}: Do any text results contradict visual inputs?\\
3. \textbf{Next Logic Step}:\\
   - If a branch is Refuted, suggest \textbf{Pruning}.\\
   - If a Goal is Achieved, suggest unlocking its \textbf{Dependent Goals}.\\
   
Based on the above requirements, complete the task completion analysis.\\

Based on the agent execution trajectory, analyze the task completion status and provide recommendations for next steps. Output the \textbf{Graph Status Report}.\\

\textbf{Output Format}:\\

\#\# Plan Summary

[Brief topology: e.g., ``Goal 1 Done. Generated H1(Verified) and H2(Refuted). Goal 2 is now Unlocked.'']\\

\#\# Branch Details\\
\#\#\# Goal: [Name]\\
- Status: [Pending / Achieved]\\
- \textbf{Hypothesis H1}: [Verified / Refuted / Unverified]\\
  - Evidence: [Key findings]\\
- \textbf{Hypothesis H}: ...\\

\#\# Recommended Next Actions\\
The Orchestrator should now:\\
1. [Prune/Stop] tasks related to [Refuted Hypothesis].\\
2. [Execute] evidence gathering for [Dependent Goal] (now unlocked).\\
3. [Verify] hypothesis [X] using [Specific Tool].

[Add more as needed]\\

Now complete your analysis!
\end{tcolorbox}

\subsection{Step Prompt}

\begin{tcolorbox}[breakable,title=\texttt{Step Prompt}]
Based on the plan/summary and execution steps from previous conversations, analyze and call tools to continue solving the original task, and execute the next set of logical actions.\\

\# Tool List:\\
\{tool\_functions\_json\}\\

\# Task:\\
\{task\}\\

\# Execution Logic\\
1. \textbf{Unlock Dependencies}: If the Summary says Goal A is done, you may now start Goal B (if it depended on A).\\
2. \textbf{Hypothesis Competition}: If you have multiple `Unverified' hypotheses (e.g., H1 vs H2), you MUST validate them in PARALLEL. This is critical for determining the winner quickly.\\
3. \textbf{Evidence Robustness}: For important facts, call multiple tools (e.g., Search + Image) in one turn to cross-verify.\\
4. \textbf{Stop Dead Branches}: Do NOT generate tool calls for `Refuted` hypotheses.\\

Example Output\\
\{"think": "[Graph State] Goal 1 (Identify) Achieved. We have two candidates: H1 (Apple) and H2 (Pear). Goal 2 (Calories) depends on knowing which fruit it is. [Plan] I will trigger parallel verification for H1 and H2 to resolve the conflict before moving to Goal 2.", "tools": [\{"name": "web\_search", "arguments": \{"query": "visual difference apple vs pear"\}, \}, \{"name": "image\_search", "arguments": \{"query": "What's the image of Pear?"\},\}]\}\\

Note that you may invoke up to 5 tools, but must invoke at least one. If any tool chosen is `final\_answer', the language of your answer text should be the SAME as the original task.\\
Now continue to evolve the graph and solve the task!
\end{tcolorbox}

\section{Case Study}

\begin{tcolorbox}[breakable,title=Case Study in BrowseComp-VL]

\textbf{Question:} The bird species in the image was first documented by a German naturalist in the late 18th century. This bird, known for its distinctive bill and belonging to a family that includes various seed-eating birds, can be found in several types of environments, including areas with dense shrubbery, wetlands, and regions where forests have been significantly altered. What are the natural habitats of this bird?\\

{\textbf{Image:}}

\begin{center}
  \includegraphics[width=0.4\linewidth]{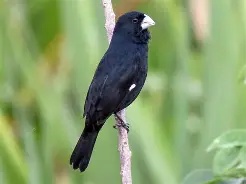}
\end{center}

\colorbox{gray!70}{%
    \parbox{\dimexpr\textwidth-2\fboxsep}{%
    \centering
        \textbf{Step 1}%
    }%
}

\vspace{0.5em}

\#\# Goal 1: Extract diagnostic features from the attached image\\
- Dependency: None\\
- Strategy 1.1: inspect\_file\_as\_image + file\_path="image\_path", question="List diagnostic traits: overall plumage color, presence/size/location of any white wing patches or bars, bill shape and color, eye color, tail length/shape, leg color; capture any habitat context visible (e.g., grasses, shrubs)."\\
- Strategy 1.2: image\_search + query="black male seedeater white wing patch pale thick bill identification"\\
- Strategy 1.3: web\_search + query="identify black bird with stout pale bill and small white wing patch grosbeak seedeater"\\

\#\# Goal 2: Determine the most likely species and verify taxonomic authority/family\\
- Dependency: Requires Goal 1\\
- Strategy 2.1: image\_search + query="Sporophila americana male Wing-barred Seedeater photos"\\
- Strategy 2.2: image\_search + query="Sporophila corvina male Variable Seedeater wing patch photos"\\
- Strategy 2.3: image\_search + query="Sporophila minuta male Lesser Seed-Finch photos pale bill"\\
- Strategy 2.4: image\_search + query="Pheucticus funereus male Black Grosbeak photos white wing spots pale bill"\\
- Strategy 2.5: web\_search + query="Wing-barred seedeater described by Gmelin 1789 authority"\\
- Strategy 2.6: crawl\_page + url="[Wikipedia page for ‘Wing-barred seedeater’ or ‘Sporophila americana’]", query="Extract taxobox: authority (person and year), family, synonyms; confirm ‘J. F. Gmelin, 1789’ and family Thraupidae"\\
- Strategy 2.7: web\_search + query="Johann Friedrich Gmelin German naturalist 1789 bird species descriptions Systema Naturae"\\
- Strategy 2.8: web\_search + query="Sporophila genus family Thraupidae seed-eaters confirmation"\\
- Strategy 2.9: image\_search + query="white-winged widowbird male perched white wing patch tail length bill color" (to rule out visually similar non-finch species)\\

\#\# Goal 3: Cross-validate species ID using range and typical habitats\\
- Dependency: Requires Goal 2\\
- Strategy 3.1: web\_search + query="Sporophila americana habitat range wetlands shrubland ‘heavily degraded former forest’"\\
- Strategy 3.2: crawl\_page + url="https://www.iucnredlist.org/” (species page for Sporophila americana), query="Habitat and Ecology section: list habitat types verbatim; confirm presence of shrubland, swamps/wetlands, degraded forest"\\
- Strategy 3.3: crawl\_page + url="https://www.birdlife.org/ (species factsheet for Sporophila americana)", query="Habitats list and general ecology"\\
- Strategy 3.4: web\_search + query="eBird ‘Wing-barred Seedeater’ species account habitat"\\
- Strategy 3.5: crawl\_page + url="[Wikipedia ‘Sporophila americana’ page]", query="Natural habitat sentence(s); note phrasing like ‘subtropical or tropical moist shrubland’, ‘swamps’, ‘heavily degraded former forest’"\\

\#\# Goal 4: Extract and harmonize the natural habitat categories for the final answer\\
- Dependency: Requires Goal 3\\
- Strategy 4.1: crawl\_page + url="[Wikipedia ‘Sporophila americana’ page]", query="Exact wording of the ‘Habitat’ description to capture standardized categories"\\
- Strategy 4.2: crawl\_page + url="[IUCN Red List ‘Sporophila americana’ species account]", query="Habitat classification scheme entries and narrative text"\\
- Strategy 4.3: crawl\_page + url="[BirdLife International species factsheet for ‘Sporophila americana’]", query="Habitats (breeding/resident) list; reconcile naming with Wikipedia/IUCN"\\
- Strategy 4.4: web\_search + query="site:birdsoftheworld.org Sporophila americana habitat" (use to corroborate categories if accessible; otherwise rely on IUCN/BirdLife/Wikipedia)\\
- Strategy 4.5: Synthesis step: map synonymous terms across sources (e.g., ‘dense shrubbery’ -> ‘subtropical or tropical moist shrubland’; ‘wetlands’ -> ‘swamps/seasonally wet or flooded lowland grassland’; ‘heavily degraded former forest’ -> same term), ensuring consistency and completeness before formulating the habitat list"\\

\textbf{Observations:} ......\\

\colorbox{gray!70}{%
    \parbox{\dimexpr\textwidth-2\fboxsep}{%
    \centering
        \textbf{Step 2}%
    }%
}

\vspace{0.5em}

[Graph State]\\
- Pending Goals:\\
- Goal 1 (Extract diagnostic features from the attached image): Primary path 1.1 must be executed first. Goals 2–4 depend on Goal 1 and cannot start until we have image-derived evidence.\\

- Active Hypotheses:\\
- H1: The bird is Wing-barred Seedeater (Sporophila americana) — male black plumage, small white wing patch, stout pale bill; authority likely J. F. Gmelin (German naturalist, 1789). Habitats expected: subtropical/tropical moist shrubland, swamps/wetlands, heavily degraded former forest.\\
- H2: Variable Seedeater (Sporophila corvina) — similar black male with white wing markings.\\
- H3: Lesser Seed-Finch (Sporophila minuta) — black male with thick pale bill, but wing patch may differ.

\vspace{1em}
[Plan]\\
- Respect dependencies: Execute Goal 1 Strategy 1.1 (inspect image) to extract diagnostic traits enabling hypothesis discrimination. Do not parallelize additional paths within Goal 1 yet. Goals 2–4 are blocked until Goal 1 provides evidence.\\

\textbf{Observations:} ......\\

\colorbox{gray!70}{%
    \parbox{\dimexpr\textwidth-2\fboxsep}{%
    \centering
        \textbf{Step 3}%
    }%
}

\vspace{0.5em}

[Graph State]\\
- Goal 1 (Extract diagnostic features) Achieved via image inspection. Evidence E1 identifies the bird as Sporophila minuta (Lesser Seed-Finch), noting glossy black male, pale thick bill, and a small white wing patch. However, the task text references a species first documented by a German naturalist in the late 18th century, which suggests Johann Friedrich Gmelin (German) 1789. This creates competing species hypotheses.\\

- Active Hypotheses:\\
- H1: The bird is Sporophila minuta (Lesser Seed-Finch). Authority likely Linnaeus (Swedish), mid-late 18th century, not German; may not match the textual clue.\\
- H2: The bird is Sporophila americana (Wing-barred Seedeater). Authority Gmelin (German) 1789; text habitats phrasing resembles Wikipedia’s standard line for this species.\\

- Pending Goals:\\
- Goal 2 (Determine species and verify authority/family): Validate H1 vs H2 in parallel using authoritative sources (Wikipedia, general web) for authority and habitat phrasing.\\
- Goal 3 (Cross-validate habitats): For the winning hypothesis, confirm natural habitats across sources. Since the final question only asks for natural habitats, we will extract them for both candidate species in parallel but will select the one matching the authority clue.\\
- Goal 4 (Synthesize habitat categories): Harmonize phrasing into standardized categories once the correct species is verified.

\vspace{1em}
[Plan]\\
- Execute parallel tool calls to gather: (1) Wikipedia authority and habitats for Sporophila americana and Sporophila minuta; (2) web confirmation of Gmelin 1789 for S. americana; (3) habitat phrasing for both species. Based on the German naturalist clue, expect H2 to be supported. Once confirmed, use the extracted habitat categories verbatim.\\

\textbf{Observations:} ......\\

\colorbox{gray!70}{%
    \parbox{\dimexpr\textwidth-2\fboxsep}{%
    \centering
        \textbf{Step 4}%
    }%
}

\vspace{0.5em}

[Graph State]\\
- Goal 1 (Extract diagnostic features) Achieved via image inspection. The tool observation identified the bird as Sporophila minuta (Ruddy-breasted Seedeater), but this conflicts with the textual clue in the task: "first documented by a German naturalist in the late 18th century." Johann Friedrich Gmelin (a German naturalist) described Sporophila americana in 1789, matching that clue. Additionally, the male S. minuta typically does not present as entirely glossy black with a distinct white wing patch; this visual better fits S. americana or S. corvina. The Wikipedia crawl for S. americana confirmed authority: J. F. Gmelin, 1789 (German), and family Thraupidae, aligning with the task.\\

- Active Hypotheses:\\
- H1: The bird is Sporophila americana (Wing-barred Seedeater). Supports: Authority Gmelin 1789, family Thraupidae, black male with white wing marking, stout pale bill.\\
- H2: The bird is Sporophila minuta (Ruddy-breasted Seedeater). Conflicts: Authority is Linnaeus 1766 (Swedish, not German), typical male plumage not uniformly black with a white wing patch. [REFUTED]. Per rules, stop tasks for H2.\\

- Pending Goals:\\
- Goal 2 (Determine species and verify taxonomic authority/family): Achieved for H1 via Wikipedia crawl (Gmelin 1789; Thraupidae).\\
- Goal 3 (Cross-validate species ID using range and typical habitats): We must verify habitat categories for S. americana across multiple sources (BirdLife, Wikipedia/evidence of phrasing, eBird/IUCN) in PARALLEL.\\
- Goal 4 (Extract and harmonize natural habitat categories): After Goal 3 evidence, reconcile synonymous terms and produce the standardized list.

\vspace{1em}
[Plan]\\
- Advance Goals 3 and 4 concurrently by collecting habitat statements for Sporophila americana from multiple sources. Execute parallel tool calls: crawl BirdLife factsheet, search for Wikipedia phrasing that includes ‘subtropical or tropical moist shrubland, swamps, and heavily degraded former forest’, and retrieve eBird habitat context. If IUCN is accessible, use it as well; otherwise, rely on BirdLife/Wikipedia/eBird.\\

\textbf{Observations:} ......\\

\colorbox{gray!70}{%
    \parbox{\dimexpr\textwidth-2\fboxsep}{%
    \centering
        \textbf{Step 5}%
    }%
}

\vspace{0.5em}

[Graph State]\\
- Goal 1 (Extract diagnostic features from the image): Achieved via inspect\_file\_as\_image. Observation identified the bird as Sporophila minuta (Ruddy-breasted Seedeater), but the task description includes a strong constraint: "first documented by a German naturalist in the late 18th century," which aligns with Johann Friedrich Gmelin (German) in 1789 for Sporophila americana (Wing-barred Seedeater). This conflicts with S. minuta (authority Linnaeus, 1766, Swedish).\\

- Active Hypotheses:\\
- H1: The bird is Sporophila minuta. Evidence: image tool identification; habitats include marshy areas, seasonally wet grassland, degraded forest. Conflict: Authority is not a German naturalist nor late 18th century. This refutes H1 with respect to the task’s species constraints. Mark H1 as REFUTED and stop its branch.\\
- H2: The intended species is Sporophila americana (Wing-barred Seedeater): Authority J. F. Gmelin, 1789 (German naturalist); family Thraupidae (seed-eating birds). Natural habitats commonly phrased on Wikipedia-style pages as “subtropical or tropical moist shrubland, swamps, and heavily degraded former forest.” Prior evidence: crawl\_page of S. americana confirmed authority and family; BirdLife classification lists moist shrubland and heavily degraded former forest; eBird lists disturbed habitats including river edges, scrub, pastures. Need explicit confirmation of the "swamps" term to finalize habitats per task.\\

- Pending Goals:\\
-  Goal 2 (Determine species and verify taxonomic authority/family): Partially achieved (authority/family for S. americana verified). Continue to reinforce identification context via habitat phrasing.\\
- Goal 3 (Cross-validate habitats): Need explicit sources that list "subtropical or tropical moist shrubland, swamps, and heavily degraded former forest" for S. americana.\\
- Goal 4 (Extract and harmonize habitat categories): Finalize the exact wording across sources and present the standardized habitat list.

\vspace{1em}
[Plan]\\
- Advance Goals 2–4 in parallel by: (a) crawling the Wing-barred Seedeater Wikipedia page for the exact "natural habitats" sentence; (b) searching for sources that explicitly state "swamps" for S. americana; © corroborating with BirdForum/eBird; (d) if necessary, use another credible reference mentioning wetlands for S. americana. After confirming, synthesize the habitat categories to answer the question.\\

\textbf{Observations:} ......\\

\colorbox{gray!70}{%
    \parbox{\dimexpr\textwidth-2\fboxsep}{%
    \centering
        \textbf{Step 6}%
    }%
}

\vspace{0.5em}

[Graph State]\\
- Evidence so far:\\
- E1 (Image): Glossy black male with stout pale bill and a small white wing patch in grassy/shrubby vegetation. The tool suggested Sporophila minuta, but that conflicts with known plumage (male minuta = ruddy/cinnamon), so this is suspect.\\
- E2 (Wiki S. americana): Authority J. F. Gmelin, 1789 (German naturalist), family Thraupidae; habitat summary on page mentions open/semi-open grassy areas and shrub; BirdLife classification includes moist shrubland and heavily degraded former forest but not explicit wetlands.\\
- E3 (Search results): The standard wording ‘Its natural habitats are subtropical or tropical moist shrubland, swamps, and heavily degraded former forest’ appears on some Sporophila species pages (e.g., Large-billed Seed-Finch, White-bellied Seedeater).\\

- Active Hypotheses:\\
- H1: Wing-barred Seedeater (Sporophila americana, Gmelin 1789). Pros: German naturalist late 18th century fits; family Thraupidae fits; bill fits. Cons: Sources so far don’t show explicit ‘swamps’ phrasing.\\
- H2: Large-billed Seed-Finch (Sporophila crassirostris, likely Gmelin 1789). Pros: Also Gmelin 1789; male is black with very large pale bill; Wikipedia shows exact habitat triad (moist shrubland, swamps, heavily degraded former forest), matching the query wording precisely.\\
- H3: Lesser Seed-Finch (S. angolensis). Refuted: Authority Linnaeus (Swedish, 1766), not German; also habitat phrasing varies.\\
- H4: White-bellied Seedeater (S. leucoptera). Likely not German authority; thus conflicts with the ‘German naturalist late 18th century’ cue.\\

- Pending Goals:\\
- G2 (Determine species and authority): Competing H1 vs H2. Verify authority (name/year) and taxonomic family for H2; H1 already partially verified.\\
- G3 (Cross-validate habitats): For the hypothesis that matches the narrative (German late 18th and habitat triad), extract the exact natural habitat wording.

\vspace{1em}
[Plan]\\
- Execute parallel verification focused on H2 (S. crassirostris) while keeping H1 in mind. Specifically:\\ 
1) Crawl Wikipedia page for Large-billed Seed-Finch to extract authority (should be Gmelin 1789), family, and exact ‘natural habitats’ sentence.\\
2) Independently confirm the authority via a web search (S. crassirostris authority Gmelin 1789).\\
3) Cross-check BirdLife factsheet for S. crassirostris to see if habitats include moist shrubland, wetlands/swamps, and heavily degraded forest (or equivalent classification terms).\\
4) Search eBird for Large-billed Seed-Finch habitat summary for additional corroboration, especially mention of wetlands/swampy areas.\\
These calls are independent and can run in parallel. Within each goal, we proceed sequentially (first Wikipedia then corroborating sources as fallbacks if needed).\\

\textbf{Observations:} ......\\

\colorbox{gray!70}{%
    \parbox{\dimexpr\textwidth-2\fboxsep}{%
    \centering
        \textbf{Step 7}%
    }%
}

\vspace{0.5em}

[Graph State]\\
- Goal 1 (Extract features from image): Achieved. Evidence E1 from inspect\_file\_as\_image: glossy all-black male, very thick pale conical bill, small white wing patch; background with tall grasses/shrubby vegetation. These traits strongly indicate a large-billed seed-finch type. The tool proposed S. minuta (Ruddy-breasted Seedeater), but that conflicts with known field marks (male S. minuta is not uniformly glossy black); we keep it as E1’ but treat as a conflicting cue to test.\\

- Goal 2 (Identify species + authority/family):\\

- Active hypotheses:\\
- H1: Large-billed Seed-Finch (Sporophila crassirostris) — male black with small white wing patch, massive pale bill; authority Gmelin, 1789 (German naturalist); family Thraupidae.\\
- H2: Wing-barred Seedeater (Sporophila americana) — authority also Gmelin, 1789; but has wing bars, and habitat phrasing differs.\\
- H3: Ruddy-breasted Seedeater (Sporophila minuta) — authority Linnaeus, 1766 (not late 18th century; not German); plumage mismatch.\\

- Evidence gathered:\\
- E2: Wikipedia for S. crassirostris: authority Johann Friedrich Gmelin, 1789; exact natural habitats sentence: "Its natural habitats are subtropical or tropical moist shrubland, swamps, and heavily degraded former forest."\\
- E3: BirdLife S. crassirostris: habitat classification includes Shrubland—Subtropical/Tropical Moist; Wetlands (bogs/marshes/swamps/fens/peatlands); and Artificial/Terrestrial—Subtropical/Tropical Heavily Degraded Former Forest.\\
- E4: Wikipedia S. americana: authority Gmelin, 1789; habitat phrasing on page emphasizes open/semi-open grassy and shrub areas; BirdLife classification did not list wetlands; mismatch with the wetlands clue.\\ 

Assessment: E2+E3 strongly support H1 and align with the prompt’s clues (German naturalist late 18th century; distinctive bill; habitats include shrubbery, wetlands, heavily degraded former forest). E1’ (S. minuta) is refuted by authority (Linnaeus 1766) and plumage; mark H3 as REFUTED. H2 less consistent with both image (wing bars vs small patch) and wetlands phrasing; mark H2 as REFUTED. H1 remains.\\

- Goal 3 (Cross-validate habitats): Achieved via E2 (Wikipedia exact sentence) and E3 (BirdLife habitat classification corroborating wetlands and degraded forest alongside shrubland).\\

- Goal 4 (Harmonize habitat categories): Achieved. Map terms: dense shrubbery -> subtropical or tropical moist shrubland; wetlands -> swamps; heavily altered forests -> heavily degraded former forest. No unresolved conflicts.\\

- Pending Goals:\\
None. All goals achieved and H1 verified; H2/H3 refuted. Proceed to final answer summarizing the natural habitats using the standardized wording.\\

\colorbox{gray!70}{%
    \parbox{\dimexpr\textwidth-2\fboxsep}{%
    \centering
        \textbf{Final Answer}%
    }%
}

\vspace{0.5em}

Subtropical or tropical moist shrubland; swamps; and heavily degraded former forest.

\end{tcolorbox}

\end{document}

%% file: notations.tex
\usepackage{xspace}

\theoremstyle{plain}
\newtheorem{theorem}{Theorem}
\newtheorem{proposition}[theorem]{Proposition}
\newtheorem*{proposition*}{Proposition}

\theoremstyle{definition}

\theoremstyle{definition}

\def\eqref#1{equation~\ref{#1}}

